\documentclass[twoside]{article}

\usepackage[accepted]{aistats2024}
\usepackage[round]{natbib}

\usepackage{amsmath}
\usepackage{amssymb}
\DeclareMathOperator{\Tr}{Tr}
\DeclareMathOperator{\logdet}{log\,det}

\usepackage{amsfonts}
\usepackage{dsfont}
\usepackage{bm}

\usepackage{algorithm}
\usepackage{algpseudocode}

\usepackage{amsthm}
\newtheorem{theorem}{Theorem}

\newtheorem{lemma}[theorem]{Lemma}
\newtheorem{definition}{Definition}[section]

\usepackage{graphicx}

\usepackage{caption}
\usepackage{subcaption}

\usepackage{hyperref} 

\usepackage{bibunits}
\defaultbibliographystyle{apalike} 
\defaultbibliography{aistats2024}

\begin{document}

\begin{bibunit}

%

%

\twocolumn[

\aistatstitle{Learning Cartesian Product Graphs with Laplacian Constraints}

\aistatsauthor{ Changhao Shi \And Gal Mishne }

\aistatsaddress{ University of California San Diego } ]

\begin{abstract}
Graph Laplacian learning, also known as network topology inference, is a problem of great interest to multiple communities.
In Gaussian graphical models (GM), graph learning amounts to endowing covariance selection with the Laplacian structure.
In graph signal processing (GSP), it is essential to infer the unobserved graph from the outputs of a filtering system.
In this paper, we study the problem of learning Cartesian product graphs under Laplacian constraints. 
The Cartesian graph product is a natural way for modeling higher-order conditional dependencies and is also the key for generalizing GSP to multi-way tensors.
We establish statistical consistency for the penalized maximum likelihood estimation (MLE) of a Cartesian product Laplacian, and propose an efficient algorithm to solve the problem.
We also extend our method for efficient joint graph learning and imputation in the presence of structural missing values.
Experiments on synthetic and real-world datasets demonstrate that our method is superior to previous GSP and GM methods.
\end{abstract}

\section{Introduction}

Graphs are powerful tools for modeling relationships among a set of entities in complex systems and have become prevalent in biology \citep{pavlopoulos2011using}, neuroscience \citep{bassett2017network}, social science \citep{borgatti2009network}, and many other scientific fields.
In machine learning and artificial intelligence, there is also a growing interest in graphs for model boosting \citep{shuman2013emerging,wu2020comprehensive,ji2021survey}.

As the graph of a system is frequently unobserved, a prominent problem in graph machine learning is how to construct a graph from available data for further use.
While ad-hoc graph construction rules (e.g. k-nearest neighbor graphs) exist in many fields, it is arguably more appealing to learn those graphs in a more principled way.
To be more specific, given a set of nodes and some nodal observations attached to them, we aim to infer their edge connectivity pattern.
In the literature, this problem is termed graph learning or network topology inference~\citep{mateos2019connecting}.

Graph learning is a central problem in GSP, the subject that generalizes traditional signal processing (SP) to non-Euclidean domains \citep{shuman2013emerging,ortega2018graph}.
In analog to traditional SP, GSP uses eigenfunctions of various graph representations, such as adjacency and Laplacian matrices, to define a graph Fourier basis that transforms graph 
signals to the spectral domain, where frequency analysis and filtering can be applied.

If assuming nodal observations should be smooth with respect to the true unseen graph (usually sparse), we can define a class of graph learning methods favoring the ``smoothness prior''.
This boils down to minimizing the total variation of nodal observations with respect to the combinatorial graph Laplacian.
Interestingly, this graph learning formulation is closely related to covariance selection in GM \citep{dempster1972covariance}.
Covariance selection aims to estimate a sparse inverse covariance matrix, or a precision matrix, from a sample covariance matrix (SCM).
Enforcing Laplacian structure on the precision matrix (and considering its pseudo-inverse) will endow covariance selection with a similar form to the smoothness prior.
The resultant problem is essentially the penalized MLE of an attractive, improper Gaussian Markov random field (IGMRF), which interests the GM community.

\begin{figure}
    \centering
    \includegraphics[width=0.4\textwidth]{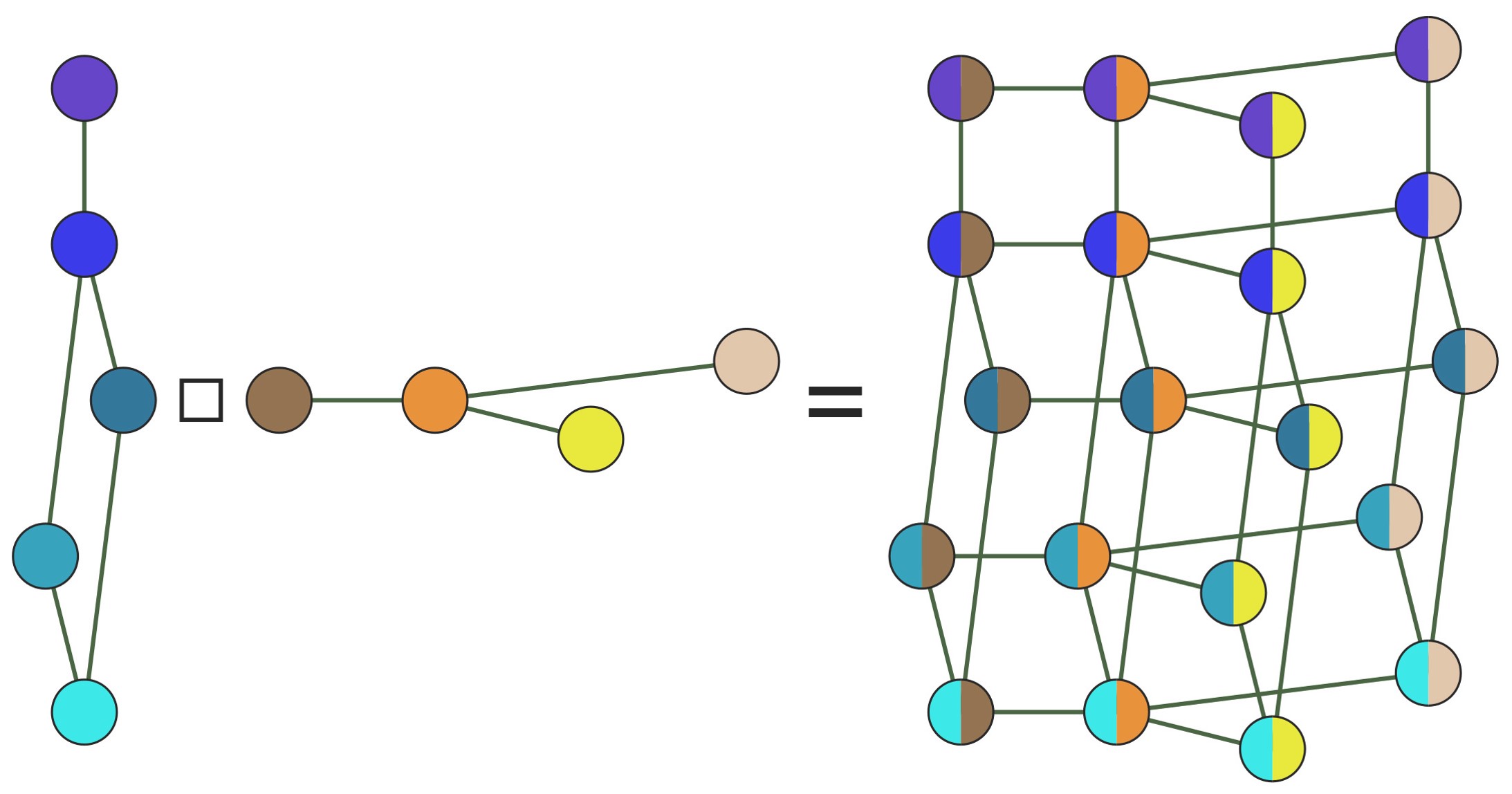}
    \caption{An example of the Cartesian graph product.}
    \label{fig:prod_example}
\end{figure}

While graph Laplacian learning and covariance selection are useful for single-way analysis, they are not intended for multi-way tensors.
A multi-way tensor, as opposed to a single-way vector, is a multi-dimensional array where each way or mode of the tensor represents a different source of variation.
Consider such a multi-way scenario: a sensor network of $p_s$ sensors with unknown connectivity and their measurements on $p_t$ time points over a day.
An example of single-way inference is to directly apply graph learning methods to learn a graph of sensors from the $p_t$ 1-d spatial signals.
Not surprisingly, this results in a sub-optimal solution since the dependencies between $p_t$ time points are ignored.
A more appealing approach is to learn a graph of size $p_sp_t$, in which each node is a (sensor, time point) pair.
However, this poses new computational challenges since $p_sp_t$ is usually huge.
To circumvent the challenges, imposing graphs with the Cartesian product structure gains massive interest. 
An example of the Cartesian graph product is shown in Fig.~\ref{fig:prod_example}.
As we can see, Cartesian product graphs are extremely suitable for multi-way data, since they offer a reasonable parsimony where only dependencies within ways are captured by factor graphs.
It is even more appealing to learn Cartesian product graphs under the Laplacian constraints, which serve as the foundations of multi-way GSP~\citep{stanley2020multiway}.
Owing to the Cartesian product Laplacian, the multi-way graph Fourier transform enjoys a concise form that separates to mode-wise Fourier transform.

In this paper, we study the problem of learning the Cartesian product Laplacian from multi-way data.
We consider the penalized MLE of the Cartesian product IGMRF and propose an efficient algorithm to solve the problem by leveraging the spectral properties of the Cartesian product Laplacian. 
A modified algorithm is also proposed for joint graph learning and missing value imputation.
On the theoretical aspect, we establish the high-dimensional statistical consistency of the proposed penalized MLE and obtain an improved rate of convergence over non-product graph Laplacian learning.
Our method provides a better solution than related GM works, which ignore the Laplacian constraints, and existing GSP works, which lack theoretical guarantees.

To summarize our contributions:
\begin{itemize}
    \item We are the first to consider the penalized MLE of Cartesian product Laplacian learning, and gain theoretical results on its asymptotic consistency, to the best of our knowledge.
    \item We propose an efficient algorithm to solve the penalized MLE, which reduces the time complexity of the naive solution. We further extend the algorithm to the setting of structural missing data. 
    \item We demonstrate that our approach outperforms existing GSP and GM methods on synthetic and real-world datasets.
\end{itemize}
As a side note, we emphasize that graph learning is intrinsically a different problem from covariance selection, although they bear a similar form.
The parameter space of graph learning and covariance selection are two disjoint sets as Laplacian matrices are singular with constant 0-eigenvectors.
Graph learning also requires that all conditional dependencies are positive, though this is also a GM subject under the study of M-matrices \citep{slawski2015estimation}.

\subsection{Related Work}
In \textbf{GSP}, the smoothness prior is arguably the most common method for graph learning \citep{dong2016learning,kalofolias2016learn,chepuri2017learning,egilmez2017graph,zhao2019optimization,buciulea2022learning}.
Other Laplacian-based models, such as heat diffusion \citep{thanou2017learning}, and models based on the normalized graph Laplacian matrix \citep{pasdeloup2017characterization} and the weighted adjacency matrix \citep{segarra2017network,navarro2020joint,shafipour2021identifying} have also been studied.

In terms of learning Cartesian product graphs, \cite{lodhi2020learning} advocated directly optimizing the total variation on product Laplacian;
\cite{Kadambari2020learning,kadambari2021product} decomposed the overall smoothness measurement into factor-wise variation so that each factor graph can be learned separately;
\cite{einizade2023learning} proposed to first estimate eigenfunctions of factor graph representations, and then solve the spectral template problem as in \citep{segarra2017network}.
However, these methods simplified the MLE to facilitate optimization, which generally leads to asymptotic inconsistencies.

In \textbf{GM}, covariance selection \citep{dempster1972covariance} aims to obtain a parsimonious model of the conditional dependencies, which amounts to learning a sparse precision matrix in a GMRF \citep{banerjee2006convex,yuan2007model,banerjee2008model}.
In modern days, this problem is efficiently solved by the prestigious graphical lasso algorithm and its variants \citep{friedman2008sparse,d2008first,rothman2008sparse,lu2009smooth,scheinberg2010sparse,li2010inexact,hsieh2011sparse,witten2011new,mazumder2012graphical,oztoprak2012newton}. 

Since then, the graphical lasso algorithm has been extended to matrix/tensor Gaussian distributions \citep{dawid1981some,gupta1999matrix} to learn Kronecker product precision matrices \citep{dutilleul1999mle,zhang2010learning,leng2012sparse,tsiligkaridis2013convergence}.
Further extensions replace the Kronecker product structure with the Kronecker sum \citep{kalaitzis2013bigraphical,greenewald2019tensor,wang2020sylvester,yoon2022eiglasso}, leading to Cartesian product graphs.
Again, none of these graphical lasso methods learn precision matrices under Laplacian constraints but only bear a similar form to the Cartesian product Laplacian learning.

\section{Background}
\label{sec:bg}

We use the following notations throughout the paper.
Lower-case and upper-case bold letters denote vectors and matrices respectively, and lower-case bold italic letters denote random vectors.
Let $\mathbf{1}$ and $\mathbf{0}$ denote the all 1 and all 0 vectors, and let $\mathbf{O}$ denote the all 0 matrix.
Let $\mathbf{e}_p^l\in\mathbb{R}^{p}$ denote a unit vector that has $1$ in its $l$-th entry. 
$\dagger$ denotes the Moore-Penrose pseudo-inverse and ${\det}^\dagger$ denotes the pseudo-determinant. 
$\circ$ denotes the Hadamard product. 
$\otimes$ and $\oplus$ denote the Kronecker product and Kronecker sum. The Kronecker sum of two matrices $\bf M_1$ and $\bf M_2$ is defined as $\bf M_1 \oplus \bf M_2= \bf M_1 \otimes \bf I_2 + \bf I_1 \otimes \bf M_2$.
For graphs $G_1$ and $G_2$, we use $\square$ to denote the Cartesian graph product operator.
Let $\times$ denote the Cartesian product of two sets.
For matrix norms, ${\|\cdot\|}_F$ denotes the Frobenius norm, ${\|\cdot\|}_2$ the operator norm,
and ${\|\cdot\|}_{1,\mathrm{off}}$ sum of the absolute values of all off-diagonal elements.
For random variables, ${\|\cdot\|}_{\psi_2}$ denotes the sub-Gaussian norm.
${(\cdot)}_+$ denotes projection to the non-negative plane and $\mathds{1}$ the indicator function.

\subsection{Preliminaries}
\label{sec:prelim}

Let $G=\{\mathcal{V},\mathcal{E},\mathbf{W}\}$ be an undirected, connected graph where $\mathcal{V}$ denotes the set of $p$ vertices, $\mathcal{E}$ the set of edges and $\mathbf{W} \in \mathcal{S}^{p}$ the weighted adjacency matrix.
Each entry of the weight matrix ${[\mathbf{W}]}_{ij}={[\mathbf{W}]}_{ji}\geq0$ encodes the similarity between a node pair $(i,j)$, and ${[\mathbf{W}]}_{ij}={[\mathbf{W}]}_{ji}>0$ iff $e_{ij}\in \mathcal{E}$.
We assume there are no self-loops, i.e. $\mathbf{W}_{ii}=0$, and denote the vectorization of all the weights as $\mathbf{w}\in\mathbb{R}^{p(p-1)/2}$, such that ${[\mathbf{w}]}_{i-j+\frac{1}{2}(j-1)(2n-j)}={[\mathbf{W}]}_{ij},\forall 1\leq j\leq i\leq p$. 
The combinatorial graph Laplacian matrix $\bf L$ of the graph $G$ is given by $\mathbf{L} =\textbf{D}-\mathbf{W}$, where $\textbf{D}$ denotes the diagonal degree matrix where ${[\mathbf{D}]}_{ii}=\sum_j {[\mathbf{W}]}_{ij}$.
$\mathbf{W}$ and $\bf L$ are often referred to as graph representations, as each can fully determine the graph $G$.

In addition, we follow \citep{kumar2020unified} to define a linear operator that maps a non-negative weight vector to the corresponding combinatorial graph Laplacian.

\begin{definition}
    Define $\mathcal{L}: \mathbb{R}^{p(p-1)/2}\to\mathbb{R}^{p\times p}, \mathbf{w}\mapsto\mathcal{L}\mathbf{w}$ as the following linear operator
    \[
    {[\mathcal{L}\mathbf{w}]}_{ij} = 
    \begin{cases}
    -{[\mathbf{w}]}_{i-j+\frac{1}{2}(j-1)(2p-j)} & i>j,\\
    {[\mathcal{L}\mathbf{w}]}_{ji} & i<j,\\
    -\sum_{k\neq j}{[\mathcal{L}\mathbf{w}]}_{kj} & i=j.
    \end{cases}
    \]
\end{definition}
One can verify that $\mathcal{L}\mathbf{w}$ is a combinatorial graph Laplacian with weights $\mathbf{w}$.
We then define its adjoint operator $\mathcal{L}^*$ such that $\langle \mathcal{L}\mathbf{w},\mathbf{Q} \rangle = \langle \mathbf{w}, \mathcal{L}^*\mathbf{Q} \rangle,\forall\mathbf{Q}\in\mathbb{R}^{p\times p}$.
\begin{definition}
    Define $\mathcal{L}^*: \mathbb{R}^{p\times p}\to\mathbb{R}^{p(p-1)/2}, \mathbf{Q}\mapsto\mathcal{L}^*\mathbf{Q}$ as the following
    \[
    {[\mathcal{L}^*\mathbf{Q}]}_l = {[\mathbf{Q}]}_{ii}-{[\mathbf{Q}]}_{ij}-{[\mathbf{Q}]}_{ji}+{[\mathbf{Q}]}_{jj},\
    \]
    \[
    l = i-j+\frac{1}{2}(j-1)(2p-j),\ i>j.
    \]
\end{definition}

Consider two weighted undirected graphs $G_1=\{\mathcal{V}_1,\mathcal{E}_1,\mathbf{W}_1\}$ and $G_2=\{\mathcal{V}_2,\mathcal{E}_2,\mathbf{W}_2\}$, with cardinality $|\mathcal{V}_1|=p_1$ and $|\mathcal{V}_2|=p_2$.
The Cartesian product of them is denoted as $G = G_1 \square G_2$, where $G_1$ and $G_2$ are referred to as the factor graphs.
The vertex set of $G$ is defined as $\mathcal{V} =\mathcal{V}_1 \times \mathcal{V}_2$.
So each node $v\in\mathcal{V}$ is indexed by a node pair $(v_1\in\mathcal{V}_1,v_2\in\mathcal{V}_2)$, and the cardinality of $G$ is $p=p_1 p_2$.
For a node pair $(v_1,v_2)$ and $(u_1,u_2)$ in the product graph $G$, $(v_1,v_2) \sim (u_1,u_2)$ holds iff $v_1=v_2 \wedge u_1 \sim u_2$ or $v_1 \sim v_2 \wedge u_1=u_2$.
The weighted adjacency matrix of $G$ is the Kronecker sum of the factor weights $\mathbf{W}=\mathbf{W}_1 \oplus \mathbf{W}_2$, and similarly for the Laplacian $\mathbf{L}=\mathbf{L}_1 \oplus \mathbf{L}_2$.

\subsection{Smoothness Prior}
\label{sec:sp}

Formally, let a graph signal be a random variable ${\bm{\mathit{f}}}:\mathcal{V} \rightarrow \mathbb{R}^p$ that assigns a real value to each vertex of the graph.
Let $\mathbf{f}$ be an instantiation of the graph signal.
The Laplacian quadratic form, also known as the Dirichlet energy of $\mathbf{f}$, is defined as $\mathbf{f}^T\mathbf{L}\mathbf{f}=\sum_{ij}{[\mathbf{W}]}_{ij}{({[\mathbf{f}]}_i-{[\mathbf{f}]}_j)}^2$, which measures the smoothness (variation) of $\mathbf{f}$ with respect to $G$.
Given $n$ graph signals (instantiations) 
and their SCM $\mathbf{S}=\frac{1}{n}\sum_{k=1}^n\mathbf{f}_k\mathbf{f}_k^T$, 
$\mathcal{J}(\{\mathbf{f}_k\}):=\Tr{(\textbf{LS})}$ measures the overall smoothness of these signals with respect to the graph.
GSP seeks to learn the Laplacian  $\mathbf{L}$ by solving the regularized smoothness problem
\begin{equation}
\label{eq:gsp_obj}
    \min_{\mathbf{L} \in \Omega_\mathbf{L}}  \left\{ \Tr(\mathbf{L} \mathbf{S}) + \alpha h(\mathbf{L}) \right\},
\end{equation}
where $\Omega_\mathbf{L}$ is the set of all combinatorial graph Laplacian matrices
\begin{equation}
\label{eq:lp_set}
    \Omega_\mathbf{L} := \left\{ \mathbf{L} \in \mathbf{S}_{+}^p \ | \ \mathbf{L}\mathbf{1}=\mathbf{0}, {[\mathbf{L}]}_{ij}={[\mathbf{L}]}_{ji} \leq 0, \forall i \neq j  \right\},\nonumber
\end{equation}
$h(\mathbf{L})$ is a regularization term, and $\alpha>0$ is a regularization parameter.
Minimizing $\mathcal{J}(\{\mathbf{f}_k\})$ encourages the signals $\{\mathbf{f}_k\}$ to vary smoothly on the inferred graph $\mathbf{L}$. 
The regularization $h(\mathbf{L})$ encodes structural priors such as sparsity, and more importantly it prevents degenerate solutions such as a set of $p$ isolated nodes resulting in  $\mathbf{L}=\mathbf{O}$.

From the perspective of GM, let an IGMRF be $\bm{\mathit{f}} \sim \mathcal{N}(\mathbf{0},\mathbf{L}^\dagger)$, 
the penalized MLE gives an estimation of $\mathbf{L}$
\begin{equation}
\label{eq:ggm_mle}
    \min_{\mathbf{L} \in \Omega_\mathbf{L}}  \left\{ \Tr(\textbf{LS}) - \log{{\det}^\dagger(\mathbf{L})} + \alpha {\|\mathbf{L}\|}_{1,\mathrm{off}} \right\}.
\end{equation}
The additional $\ell_1$ regularization promotes sparsity, and $\alpha>0$ controls its strength.
The penalized MLE is almost a standard covariance selection problem, with the only difference being that the precision matrix is constrained to be a combinatorial graph Laplacian.

We notice the similarity between \eqref{eq:gsp_obj} and \eqref{eq:ggm_mle}.
The IGMRF formulation \eqref{eq:ggm_mle} can be interpreted as a particular case of the GSP formulation \eqref{eq:gsp_obj}, where $h(\mathbf{L})=- \log{{\det}^\dagger(\mathbf{L})} + \alpha {\|\mathbf{L}\|}_{1,\mathrm{off}}$.
To further see the connection from IGMRF to GSP, consider a system $\bm{\mathit{f}}=\mathcal{F}(\mathbf{L})\bm{\mathit{f}}_0$ where $\mathcal{F}(\mathbf{L})$ is the graph filter.
Let the input signals be random Gaussian noise $\bm{\mathit{f}}_0 \sim \mathcal{N}(\mathbf{0},\mathbf{I}_p)$, and let the graph filter be a low-pass one $\mathcal{F}(\mathbf{L})=\sqrt{\mathbf{L}^\dagger}=\textbf{U}\sqrt{\boldsymbol\Lambda^\dagger}\textbf{U}^T$, where $\mathbf{L}=\textbf{U}\boldsymbol\Lambda\textbf{U}^T$ is the eigendecomposition.
This leads to the same $\bm{\mathit{f}} \sim \mathcal{N}(\mathbf{0},\mathbf{L}^\dagger)$.
Thus, fitting this IGMRF is equivalent to estimating the graph filter of the given form under previous assumptions.

\section{Product Graph Learning}
\label{sec:pgl}

\subsection{Penalized MLE}
\label{sec:pl_problem}

Let the random matrix $\bm{\mathit{X}}\in\mathbb{R}^{p_1\times p_2}$ represent a two-way graph signal that lives on the product graph $G$. 
${[\bm{\mathit{X}}]}_{i_1,i_2}$ is the signal on node $(i_1,i_2)$.
Given $n$ instantiations $\{\textbf{X}_1,\textbf{X}_2,\dots,\textbf{X}_n\}$, our goal is to learn the factor graphs $G_1, G_2$ and their Cartesian product $G$ from these nodal observations on $G$.
Note that to ease the presentation, we limit the number of factor graphs to two, but our formulation and solution can be easily generalized to more factors and higher-dimensional tensors $\bm{\mathit{X}}\in\mathbb{R}^{p_1\times p_2\times p_3 \times \dots}$.

Let the random vector $\bm{\mathit{x}}$ be the vectorization of $\bm{\mathit{X}}$ and $\mathbf{S}=\frac{1}{n}\sum_{k=1}^{n}\textbf{x}_k{\textbf{x}_k}^T$ be the SCM.
Since for $G = G_1 \square G_2$ we have $\mathbf{L}=\mathbf{L}_1\oplus\mathbf{L}_2$ \citep{barik2018spectra}, we derive the product graph learning objective 
\begin{equation}
\label{eq:pggm_mle}
\begin{gathered}
    \min_{\mathbf{L}_1,\mathbf{L}_2 \in \Omega_\mathbf{L}}  \Bigl\{ \Tr((\mathbf{L}_1\oplus\mathbf{L}_2) \mathbf{S}) - \log{{\det}^\dagger(\mathbf{L}_1\oplus\mathbf{L}_2)}\\ + \alpha {\|\mathbf{L}_1\oplus\mathbf{L}_2\|}_{1,\mathrm{off}} \Bigr\}.
\end{gathered}
\end{equation}
Similar to \eqref{eq:ggm_mle}, \eqref{eq:pggm_mle} can be interpreted from either a GSP or a GM perspective, which we discuss below.

\paragraph{GSP Interpretation:}
We decompose $\mathcal{J}(\{\mathbf{X}_k\})=\Tr((\mathbf{L}_1\oplus\mathbf{L}_2)\mathbf{S})=\Tr{(\mathbf{L}_1\mathbf{S}_1)}+\Tr{(\mathbf{L}_2\mathbf{S}_2)}$, where
\begin{align}
    & \mathbf{S}_1=\frac{1}{n}\sum_{k=1}^n\mathbf{X_k}{\mathbf{X_k}}^T,
    & \mathbf{S}_2=\frac{1}{n}\sum_{k=1}^n{\mathbf{X_k}}^T\mathbf{X_k}.
    \label{eq:s1s2}
\end{align}
This indicates that the variation on the product graph equals the sum of mode-wise variation, and Cartesian product graph learning encourages signals to be smooth on each factor.
In contrast to existing non-consistent GSP methods, we use a log-determinant regularization naturally induced by MLE, which is crucial for the estimator to be consistent as we will show.

\paragraph{GM Interpretation:}
Consider a random matrix-variate $\bm{\mathit{X}}$ defined by a IGMRF $\bm{\mathit{x}} \sim \mathcal{N}(\mathbf{0},{(\mathbf{L}_1\oplus\mathbf{L}_2)}^\dagger)$.
Then \eqref{eq:pggm_mle} is the penalized MLE of fitting this model to the product graph signals.
Solving \eqref{eq:pggm_mle} amounts to enforcing the Laplacian structure on the Kronecker sum precision matrices.
As the Laplacian constraints are essentially a structural prior, they are important for accurate estimation, especially when $n$ is small.
We will show that our experiments verify this claim.

\subsection{Multi-Way Graph Learning} 
\label{sec:pgd}

We now propose the \textbf{M}ulti-\textbf{W}ay \textbf{G}raph (Laplacian) \textbf{L}earning (\textbf{MWGL}) algorithm for solving \eqref{eq:pggm_mle}.
First we rewrite \eqref{eq:pggm_mle} as
\begin{align}
\label{eq:mle_pgd}
        \min_{\mathbf{w}_1,\mathbf{w}_2 \geq \mathbf{0}}  \Bigl\{ \mathbf{w}_1^T\mathcal{L}^*\mathbf{S}_1  & + \mathbf{w}_2^T\mathcal{L}^*\mathbf{S}_2 - \log{{\det}^\dagger(\mathcal{L}\mathbf{w}_1\oplus\mathcal{L}\mathbf{w}_2)} \nonumber \\ &+ \alpha_1\mathbf{w}_1^T\mathbf{1} + \alpha_2\mathbf{w}_2^T\mathbf{1} \Bigr\},
\end{align}
since $\Tr(\mathbf{LS})=\Tr(\mathcal{L}\mathbf{w}\mathbf{S})=\mathbf{w}^T\mathcal{L}^*\mathbf{S}$.
The absolute sign of the $\ell_1$ norm of the sparsity regularization is redundant due to the non-negative constraints.
We then use projected gradient descent to solve $\mathbf{w}_1$ and $\mathbf{w}_2$.
The update of $\mathbf{w}_1$ and $\mathbf{w}_2$ is given by
\begin{equation}
    \begin{gathered}
    \mathbf{w}_1^{(t+1)} = {(\mathbf{w}_1^{(t)}-\eta(\mathcal{L}^*\mathbf{S}_1-\mathcal{L}^*\mathbf{H}_1^{(t)}+\alpha_1\mathbf{1}))}_+,\\
    \mathbf{w}_2^{(t+1)} = {(\mathbf{w}_2^{(t)}-\eta(\mathcal{L}^*\mathbf{S}_2-\mathcal{L}^*\mathbf{H}_2^{(t)}+\alpha_2\mathbf{1}))}_+,
    \end{gathered}\label{eq:w1w2_update}
\end{equation}
where 
for $\mathbf{H}_1\in \mathbb{R}^{p_1\times p_1}$ and $\mathbf{H}_2\in \mathbb{R}^{p_2\times p_2}$ we have
\begin{equation}
    \begin{gathered}
    \mathbf{H}_1 = \sum_{l=1}^{p_2} {(\mathbf{I}_{p_1}\otimes \mathbf{e}_{p_2}^l)}^T {(\mathcal{L}\mathbf{w}_1\oplus\mathcal{L}\mathbf{w}_2)}^\dagger (\mathbf{I}_{p_1}\otimes \mathbf{e}_{p_2}^l), \\
    \mathbf{H}_2 = \sum_{l=1}^{p_1} {(\mathbf{e}_{p_1}^l\otimes \mathbf{I}_{p_2})}^T {(\mathcal{L}\mathbf{w}_1\oplus\mathcal{L}\mathbf{w}_2)}^\dagger (\mathbf{e}_{p_1}^l\otimes \mathbf{I}_{p_2}).
    \end{gathered}\label{eq:h1h2_costly}
\end{equation}
The regularization parameters for each factor graph are $\alpha_1=p_2\alpha$ and $\alpha_2=p_1\alpha$, but in practice, we can benefit from a free grid search of $\alpha_1$ and $\alpha_2$.
With the learning rate $\eta$ of the user's choice, alternating between \eqref{eq:w1w2_update} until stopping criteria is satisfied solves the Cartesian product graph learning problem.
The above projected gradient descent scheme is guaranteed to converge in $\mathcal{O}(\frac{1}{t})$ for $\eta$ that is sufficiently small.

A closer look reveals that this solution is not computationally scalable.
Computing the gradient involves taking the inverse of the product graph Laplacian, which can be huge when the number of factors increases.
Even for the 2-factor case, the computational cost of inversion will explode quickly as the size of each factor graph grows.
Fortunately, we can compute $\mathbf{H}_1$ and $\mathbf{H}_2$ efficiently using the following lemma.
\begin{lemma}[Efficient Computation]
\label{lm:efficient_h1h2}
    The $\mathbf{H}_1$ and $\mathbf{H}_2$ matrices defined in \eqref{eq:h1h2_costly} can be efficiently computed as
    \begin{equation}
        \begin{gathered}
        \mathbf{H}_1 = \mathbf{U}_1 \sum_{l=1}^{p_2} {(\mathbf{\Lambda}_1+{[\bm{\Lambda}_2]}_{l,l}\mathbf{I}_{p_1})}^\dagger \mathbf{U}_1^T,\\
        \mathbf{H}_2 = \mathbf{U}_2 \sum_{l=1}^{p_1} {(\mathbf{\Lambda}_2+{[\bm{\Lambda}_1]}_{l,l}\mathbf{I}_{p_2})}^\dagger \mathbf{U}_2^T,
        \end{gathered}\label{eq:h1h2}
    \end{equation}
    where $\mathbf{L}_1=\mathbf{U}_1\boldsymbol\Lambda_1\mathbf{U}_1^T$ and $\mathbf{L}_2=\mathbf{U}_2\boldsymbol\Lambda_2\mathbf{U}_2^T$
    are the eigendecompositions of factor Laplacian matrices.
\end{lemma}
The key to obtaining \eqref{eq:h1h2} is to leverage the spectral structure of the Cartesian product Laplacian and its inverse.
A similar strategy has been used in \citep{yoon2022eiglasso}, but here we do not suffer from the identifiability issue thanks to the Laplacian constraints.
See the supplement for the details.
Note that \eqref{eq:h1h2} only requires matrix operations on the factor scales and avoids taking the cumbersome inversion of $\mathbf{L}$ as in \eqref{eq:h1h2_costly}.
These two steps together demand $\mathcal{O}(p_1^3+p_2^3)$ time complexity (dominated by eigendecompositions), which is a significant reduction from the full matrix inversion $\mathcal{O}(p^3)=\mathcal{O}(p_1^3p_2^3)$.
Alg.~\ref{alg:cpgl} summarizes the algorithm.
\begin{algorithm}
\caption{MWGL}
\label{alg:cpgl}
\textbf{Input:} graph signals $\{\mathbf{X}_k\}$, parameters $\alpha,\eta$
\begin{algorithmic}
\State Compute $\mathbf{S}_1,\mathbf{S}_2$ as in \eqref{eq:s1s2}.
\State Initialize $\mathbf{w}_1$ and $\mathbf{w}_2$.
\Repeat
    \State Compute $\mathbf{H}_1$ and $\mathbf{H}_2$ as in \eqref{eq:h1h2}.
    \State Update $\mathbf{w}_1$ and $\mathbf{w}_2$ with \eqref{eq:w1w2_update}.
\Until{convergence or reaching maximum iterations.}
\end{algorithmic}
\textbf{Output:} factor graph weights $\mathbf{w}_1,\mathbf{w}_2$
\end{algorithm}

\subsection{Structural Missing Values}
\label{sec:mv}

Missing values are common in real-world data.
In some cases, there are random missing entries in $\mathbf{X}_k$;
in other cases, the entire fiber $\{{[\mathbf{x}_1]}_i,{[\mathbf{x}_2]}_i,\dots,{[\mathbf{x}_n]}_i\}$ of node $i$
is missing.
Inferring connectivity of these missing nodes is generally impossible unless the underlying graph is a Cartesian product.
An example, which we demonstrate in the experiments, is learning the product graph from multi-view object images when images of some (object, view) pairs are not accessible.
For these missing nodes, their object edges are preserved by other views of the same object, and their view edges are preserved by other objects of the same view.

We now propose to learn the graphs and impute the missing values simultaneously.
Let $\Psi^\complement$ be the set of missing nodes in the product graph.
We treat missing values as contamination of the true data and refine the imputation before every projected gradient descent step in Alg.~\ref{alg:cpgl}, i.e.  we alternate between filling in the data and learning the factor graphs as before.
Let $\mathbf{X}_k^{(t)}$ be the imputed signals at step $t$ and the signals on the observed nodes ${[\mathbf{X}_k^{(t)}]}_{\Psi} ={[\mathbf{X}_k]}_{\Psi}$ are fixed.
Consider 
\begin{equation}
\label{eq:impute}
        \min_{\{\mathbf{X}_k^{(t)}\}}  \Bigl\{ \frac{1}{2n\beta}\sum_{k=1}^n{\|\mathbf{X}_k^{(t-1)}-\mathbf{X}_k^{(t)}\|}_F^2 + \mathcal{J}(\{\mathbf{X}_k^{(t)}\}) \Bigr\},
\end{equation}
where $\beta$ is a trade-off parameter.
We solve $\{{[\mathbf{X}_k^{(t)}]}_{\Psi^\complement}\}$ inexactly by alternating the following steps
\begin{equation}
    \begin{gathered}
    {[\widetilde{\mathbf{X}}_k^{(t-1)}]}_{\Psi^\complement} = {[{(\beta\mathbf{L}_1^{(t-1)}+\mathbf{I}_{p_1})}^{-1} \mathbf{X}_k^{(t-1)}]}_{\Psi^\complement},\\
    {[\mathbf{X}_k^{(t)}]}_{\Psi^\complement} = {[\widetilde{\mathbf{X}}_k^{(t-1)} {(\beta\mathbf{L}_2^{(t-1)}+\mathbf{I}_{p_2})}^{-1}]}_{\Psi^\complement}.
    \end{gathered}\label{eq:tikh_l1l2}
\end{equation}
\eqref{eq:tikh_l1l2} is the partial solution of the Tikhonov filtering
\begin{equation}
    \begin{gathered}
    \min_{\{\widetilde{\mathbf{X}}_k\}} \frac{1}{2n\beta}\sum_{k=1}^n{\|\mathbf{X}_k^{(t-1)}-\widetilde{\mathbf{X}}_k\|}_F^2 + \Tr(\mathbf{L}_1\mathbf{S}_1),\\
    \min_{\{\mathbf{X}_k\}} \frac{1}{2n\beta}\sum_{k=1}^n{\|\widetilde{\mathbf{X}}_k^{(t-1)}-\mathbf{X}_k\|}_F^2 + \Tr(\mathbf{L}_2\mathbf{S}_2),
    \end{gathered}
\end{equation}
which are low-pass graph filters that smooth missing value imputations with current factor graph estimation. 
Note that we alternately filter the signals with factor graphs rather than employ one-pass filtering with the product graph.
This eases the computation for the same reason as we stated in Sec.~\ref{sec:pgd}.
We term this algorithm \textbf{MWGL-Missing} and summarize it in Alg.~\ref{alg:cpgl_mv}.
\begin{algorithm}
\caption{MWGL-Missing}
\label{alg:cpgl_mv}
\textbf{Input:} observed nodes $\Psi,\{{[\mathbf{X}_k]}_{\Psi}\}$, parameters $\alpha,\beta,\eta$
\begin{algorithmic}
\State Initialize $\mathbf{w}_1$, $\mathbf{w}_2$.
\Repeat
    \State Refine imputed values $\{{[\mathbf{X}_k]}_{\Psi^\complement}\}$ with \eqref{eq:tikh_l1l2}.
    \State Update $\mathbf{S}_1,\mathbf{S}_2$ as in \eqref{eq:s1s2}.
    \State Compute $\mathbf{H}_1$ and $\mathbf{H}_2$ as in \eqref{eq:h1h2}.
    \State Update $\mathbf{w}_1$ and $\mathbf{w}_2$ with \eqref{eq:w1w2_update}.
\Until{convergence or reaching maximum iterations.}
\end{algorithmic}
\textbf{Output:} factors $\mathbf{w}_1,\mathbf{w}_2$, imputed values$\{{[\mathbf{X}_k]}_{\Psi^\complement}\}$
\end{algorithm}

\section{Theoretical Results}

Now we establish the statistical consistency and convergence rates for the Cartesian product Laplacian estimator as in \eqref{eq:pggm_mle}.
We first make two assumptions regarding the true underlying graph we were to estimate:
\begin{itemize}
    \item[(A1)] Let $\mathcal{A}={\{(i,j)|{[\mathbf{w}]}_{i-j+\frac{1}{2}(j-1)(2p-j)} > 0,i>j\}}$ be the support set of $\mathbf{w}$. We assume the graph is sparse and the cardinality of the support set is upper bounded by $|\mathcal{A}| \leq s$.
    \item[(A2)] Let $\{0,\lambda_2,\dots,\lambda_p\}$ be the eigenvalues of the true product Laplacian in a non-decreasing order.
    We assume these eigenvalues are bounded away from $0$ and $\infty$ by a constant $z>1$, such that $\frac{1}{z} \leq\lambda_2<\lambda_p\leq z$.
\end{itemize}
Both assumptions are common in high-dimensional statistics literature.
Also notice that in our case, bounding the Fiedler value (the second smallest eigenvalue) away from $0$ implies that the graph is connected.

\begin{theorem}[Existence of MLE]
\label{th:existence}
    The penalized negative log-likelihood of Cartesian product Laplacian learning as in \eqref{eq:pggm_mle} is lower-bounded, and there exists at least one global minimizer as the solution of the penalized MLE.
\end{theorem}
\cite{ying2021minimax} proved that the negative log-likelihood as in \eqref{eq:ggm_mle} is lower-bounded and the MLE exists.
Since the Laplacian of Cartesian product graphs form a subset of all graph Laplacians, the same lower bound applies here.
In fact, we derive a tighter lower bound for the Cartesian product graphs.
It remains to show that the global minimizer can be achieved in this subspace of Cartesian product graphs.
We demonstrate this by parameterizing the product Laplacian in \eqref{eq:pggm_mle} with $\mathbf{w}_1$ and $\mathbf{w}_2$.

\begin{theorem}[Uniqueness of MLE]
\label{th:uniqueness}
    The objective function of penalized MLE is jointly convex with respect to the factor graphs, and its global minimizer uniquely exists.
\end{theorem}
The uniqueness is not surprising since the original graph Laplacian learning problem is convex, and the map from factor graphs to their Cartesian product is linear.

\begin{theorem}[High-dimensional consistency]
\label{th:consistency}
    Suppose assumptions (A1) and (A2) hold.
    Then with sufficient observations
    \begin{equation}
        n \geq \max \left[ \frac{c^2s\log p}{\lambda^2_p\min(p_1,p_2)}, \frac{c_2^2\log p}{64\min(p_1,p_2)} \right],
    \end{equation}
    and regularization parameter
    \begin{equation}
        \alpha = \frac{c_2}{2\lambda_2}\sqrt{\frac{\log p}{n\min(p_1,p_2)}},
    \end{equation}
    the minimizer $\widehat{\mathbf{L}}$ of the penalized MLE as in \eqref{eq:pggm_mle} is asymptotically consistent to the true Laplacian $\mathbf{L}^*$ with the Frobenius error bound
    \begin{equation}
    \label{ieq:convergence_rate}
        {\|\widehat{\mathbf{L}}-\mathbf{L}^*\|}_F \leq c \sqrt{\frac{\log p}{n\min(p_1,p_2)}},
    \end{equation}
    in probability $1-2\exp{(-c'\log p)}$. 
    $c_1$, $c_2$, $c>\frac{8\sqrt{2}c_2\lambda^2_p}{\lambda_2}$, and $c'=\frac{c_1c_2^2}{64}-2$ are constants.
\end{theorem}
Theorem~\ref{th:consistency} not only proves that our proposed estimator is guaranteed to converge to the true Laplacian but also improves the rate of consistency from \citep{ying2021minimax} by a factor of $\sqrt{\min(p_1,p_2)}$.
The improvement reflects the recurrence of factor dependencies in a single product graph signal.
A similar trend has been observed in \citep{greenewald2019tensor} when the graphical lasso generalizes to multi-dimensional tensors.
The key to proving the improved rate is using Hanson-Wright inequality \citep{hanson1971bound,rudelson2013hanson} to obtain concentration results on individual modes of the multi-way tensor.
For $\min(p_1,p_2)=1$, our convergence rate coincides with the one in \cite{ying2021minimax}.
See detailed proofs of the above theorems in the supplement.

\section{Experiments}
\label{sec:exp}

We conduct extensive experiments in MATLAB on both synthetic and real-world datasets to evaluate our method.
See the supplement for more details.

\subsection{Synthetic Graphs}
\label{sec:exp_synthetic}

\begin{figure*}[hbt!]
\centering
\includegraphics[width=\textwidth]{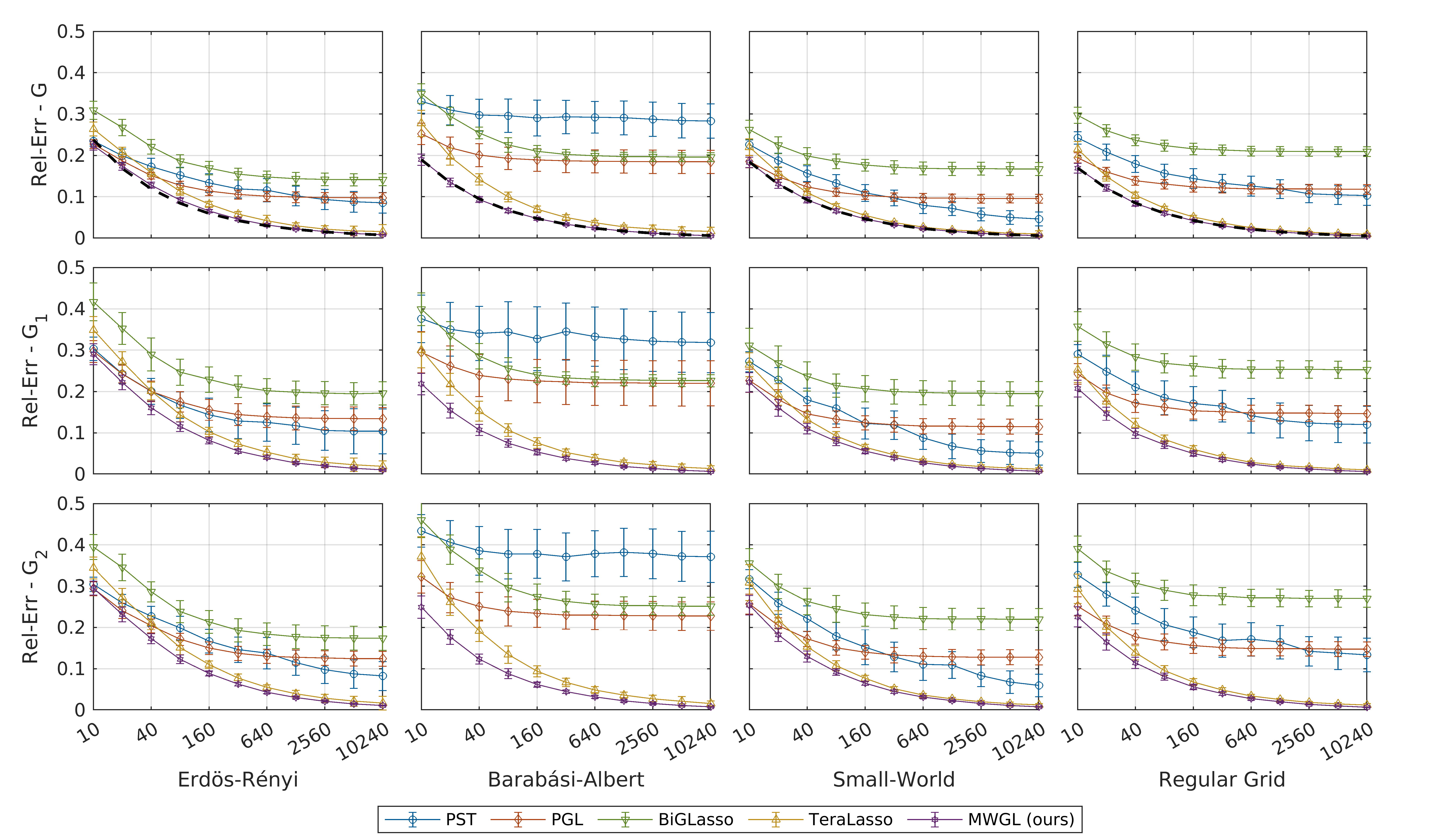}
\caption{Comparison of different methods on various synthetic data.
Each sub-figure shows the trend of Rel-Err of the product or factor Laplacian matrices as $n$ increases.
Black dash lines fit the theory in \eqref{ieq:convergence_rate} to our results.}
\label{fig:re}
\end{figure*}

We first evaluate on synthetic graphs and signals.
We use the following models to generate factor graphs: 
\begin{enumerate}
    \item[(1)] Erd\H{o}s-R\'enyi model with edge probability $0.3$;
    \item[(2)] Barab\'asi-Albert model with preferential attachment $2$ starting from $2$ initial nodes;
    \item[(3)] Watts-Strogatz small-world model, where we create an initial ring lattice with degree $2$ and rewire every edge of the graph with probability $0.1$;
    \item[(4)] and regular grid model.
\end{enumerate}
Edge weights are then randomly sampled from a uniform distribution $\mathcal{U}(0.1,2)$ for each edge.
We generate weighted factor graphs of $p_1=20$ and $p_2=25$ nodes using each graph model and take their Cartesian product to obtain graphs of $p=p_1p_2=500$ nodes.
The factor grid models are $4\times5$ grids and $5\times5$ grids.
The signals are then generated from $\bm{\mathit{f}}\sim \mathcal{N}(\mathbf{0},\mathbf{L}^{\dagger})$.
We generate $n=10\times2\string^\{0,1,\dots,10\}$ signals for each synthetic product graph, and evaluate graph learning methods under these different settings.
We repeat this process to obtain 50 realizations for each graph model. 

We compare MWGL with multiple GSP and GM baselines.
For the GSP baselines, we compare with the PGL (Product Graph Learning) method \citep{kadambari2021product} and the PST (Product Spectral Template) method \citep{einizade2023learning}.
To compare with GM methods, we select the BiGLasso \citep{kalaitzis2013bigraphical} and TeraLasso \citep{greenewald2019tensor} algorithms that learn precision matrices of Kronecker sum structure.
Since a precision matrix $\mathbf{\Theta}$ learned GM methods is generally not a Laplacian, we select its positive ``edges'' $\mathbf{w}_{\mathbf{\Theta}}={(-\mathrm{tril}(\mathbf{\Theta},-1))}_+$ and build a true Laplacian $\mathcal{L}\mathbf{w}_{\mathbf{\Theta}}$ for evaluation.
$\mathrm{tril}$ stands for the Matlab operation of lower triangular vectorization.

We use the relative error (Rel-Err) and the area under the precision-recall curve (PR-AUC) as main evaluation metrics.
Since the selected GSP baselines impose the constraints $\Tr(\mathbf{L}_1)=p_1$ and $\Tr(\mathbf{L}_2)=p_2$ (thus $\Tr(\mathbf{L})=\Tr(\mathbf{L_1}\otimes \mathbf{I}_{p_2})+\Tr(\mathbf{I}_{p_1}\otimes \mathbf{L_2})=2p_1p_2$), we normalize the true Laplacian and the Laplacian learned by other methods for the comparison of Rel-Err
\begin{equation}
    \mathbf{Ln}_1 = \frac{p_1}{\Tr(\mathbf{L}_1)}\mathbf{L}_1,\mathbf{Ln}_2 = \frac{p_2}{\Tr(\mathbf{L}_2)}\mathbf{L}_2,\mathbf{Ln} = \frac{2p_1p_2}{\Tr(\mathbf{L})}\mathbf{L}.\nonumber
\end{equation}
The Rel-Err between the true and learned Laplacian in terms of the Frobenius norm is then computed as
\begin{equation}
    \frac{{\|\widehat{\mathbf{Ln}}-\mathbf{Ln}^*\|}_F}{{\|\mathbf{Ln}^*\|}_F},
\end{equation}
similarly for the factor graphs.
We perform grid search to decide the best regularization parameter of each method.
Fig.~\ref{fig:re} shows the averaged Rel-Err across 50 realizations and the standard deviations on each setting.
We leave the PR-AUC results to the supplement.

Fig.~\ref{fig:re} demonstrates that our MWGL outperforms both GSP and GM baselines in all the settings.
PGL performs well in the low data regime but loses its advantage when $n$ increases.
The plots indicate that PGL is inherently asymptotically inconsistent, which is reasonable since their objective function misses the integral log-determinant term of the MLE.
PST improves fast as $n$ grows, since the estimated spectral template, i.e. Laplacian eigenvectors, becomes increasingly accurate.
However, it still underperforms MWGL even when $n$ is large.
For the GM baselines, TeraLasso outperforms BiGLasso and comes close to MWGL for large $n$. 
But when $n$ is small, it underperforms MWGL and sometimes other GSP baselines, which shows the importance of the Laplacian constraints as a structural prior.
Compared to all these baselines, our MWGL performs well in the full spectrum of $n$.
Also note that the Rel-Err curves of our method fit convergence rate in \eqref{ieq:convergence_rate} very well (we solve for $c$ via regression), which validates our theoretical findings.

\subsection{Molene Weather Data}
\label{sec:exp_molene}

We next consider the Molene weather dataset \citep{loukas2019stationary}, originally published by the French National Meteorological Service.
The dataset contains hourly temperature recordings of 32 weather stations in Brest, France, during the month of January 2014.
Our goal is to learn the product of a 32-node geographical graph of weather stations and a 24-node temporal graph of hours.
The daily recordings of all stations form a graph signal, and we aim to learn the Cartesian product graph from the 31 daily signals.

\begin{figure}[hbt!]
    \centering
    \begin{subfigure}[b]{0.22\textwidth}
        \centering
        \includegraphics[width=\textwidth]{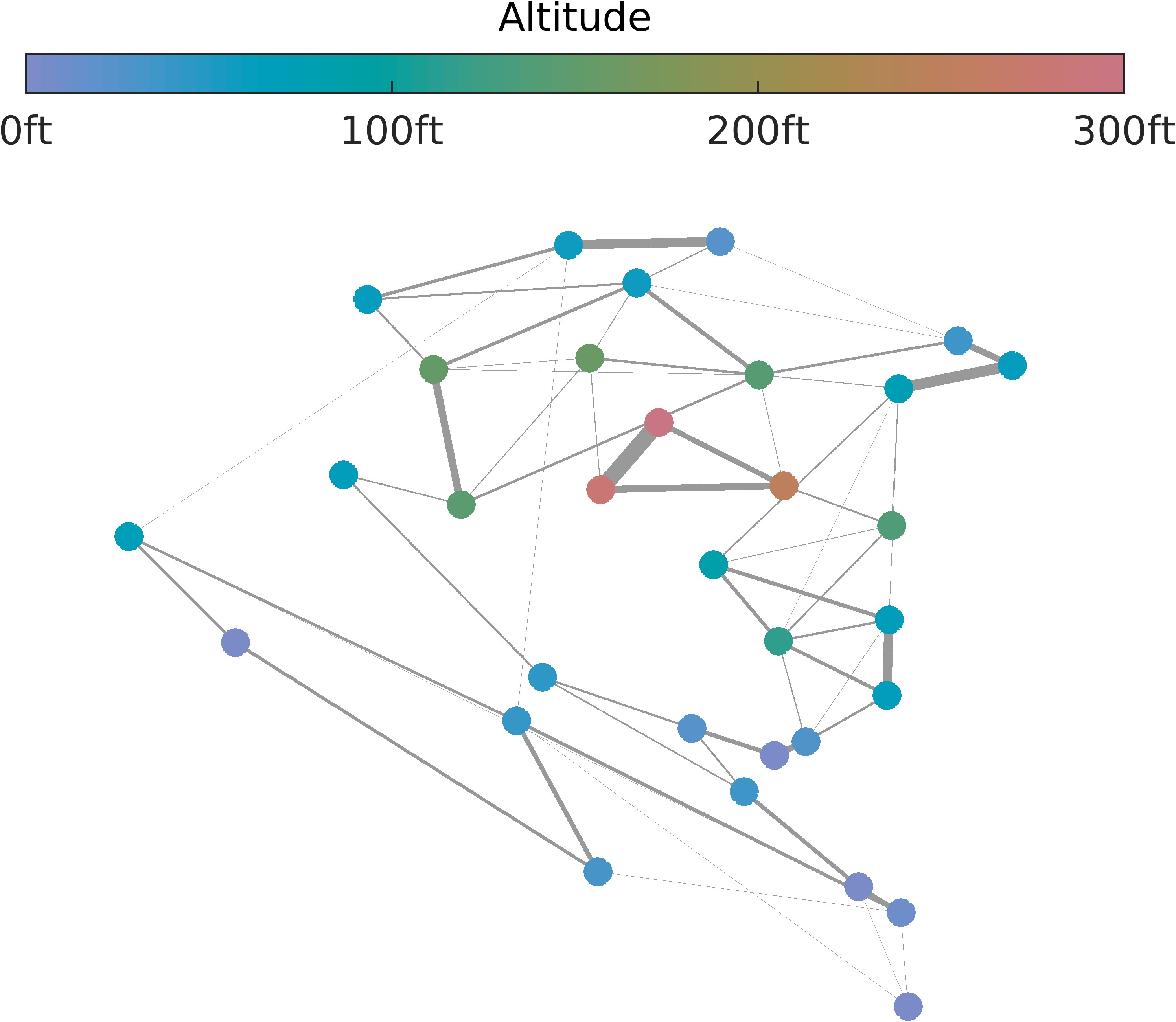}
        \caption{Stations}
    \end{subfigure}
    \hfill
    \begin{subfigure}[b]{0.227\textwidth}
        \centering
        \includegraphics[width=\textwidth]{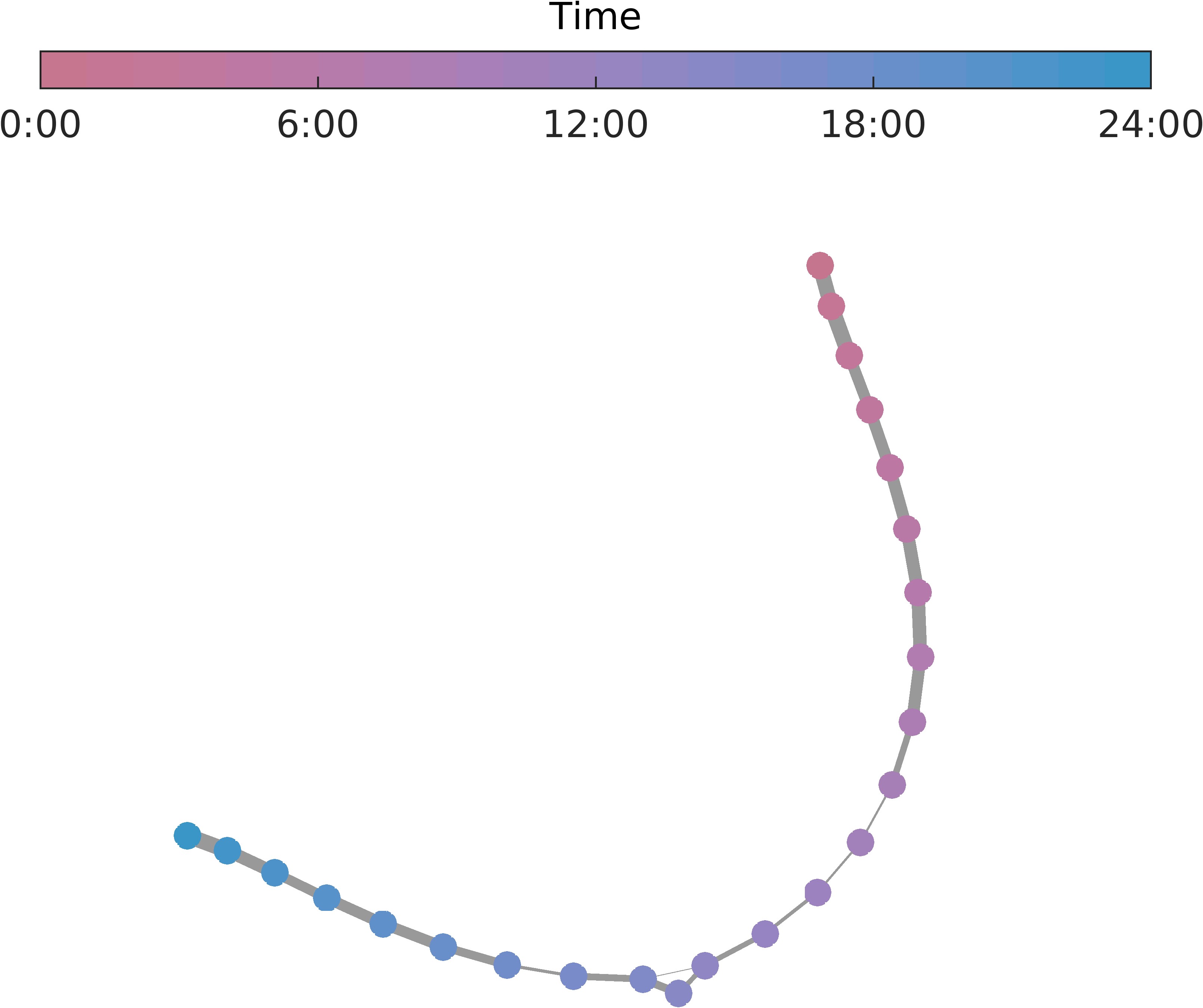}
        \caption{Hours}
    \end{subfigure}
    \caption{The inferred factor graphs of Molene. Stations are placed according to their real coordinates. 
    }
    \label{fig:molene_graphs}
\end{figure}

MWGL again learns reasonable factors as demonstrated in Fig.~\ref{fig:molene_graphs}.
The learned weather station graph faithfully reflects their coordinates and altitudes.
The 24 nodes of hours form a path graph, in alignment with their temporal order.

\subsection{COIL-20 Dataset}
\label{sec:exp_coil}

We now evaluate MWGL-Missing on the Columbia Object Image Library 20 (COIL-20) dataset \citep{nene1996columbia}.
COIL-20 consists of $128\times 128$ grey-scale images of 20 small objects, where each object is placed on a turning table and captured by a fixed camera to obtain multi-views at evenly distributed angles.
Images are taken every 5 degrees to produce 72 views per object, which we sub-sample to 36 views.
Our goal is to learn a Cartesian product of a 20-node object graph and a 36-node view graph from the $16384=128\times 128$ graph signals.
To create structural missingness, we remove the images from 180 to 360 degree of half of the objects (25$\%$ of all data) and apply MWGL-Missing.

\begin{figure}[hbt!]
    \centering
    \begin{subfigure}[b]{0.45\textwidth}
        \centering
        \includegraphics[width=\textwidth]{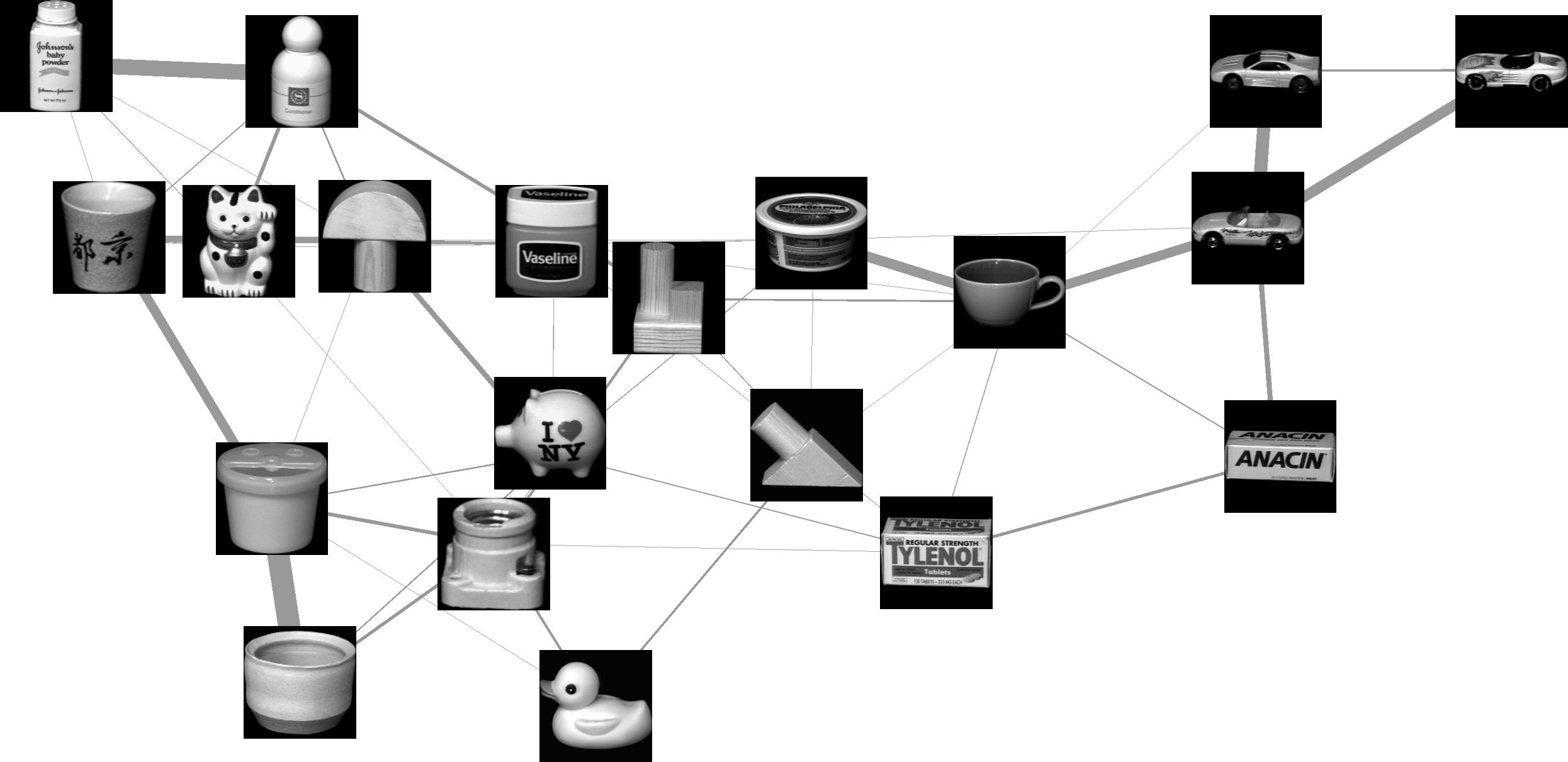}
        \caption{Objects}
    \end{subfigure}%
    \vspace{5pt}
    \begin{subfigure}[b]{0.46\textwidth}
        \centering
        \includegraphics[width=\textwidth]{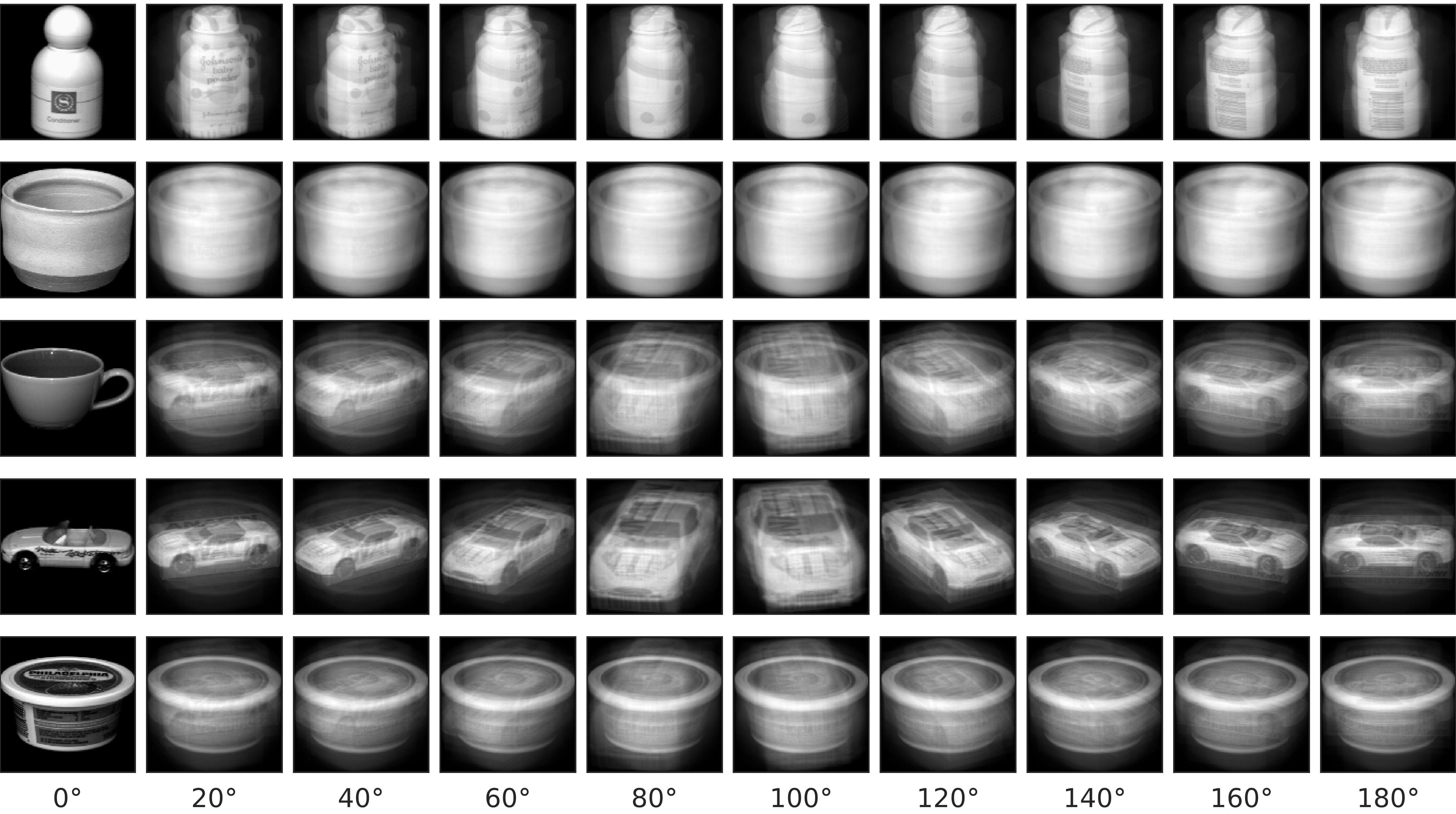}
        \caption{Imputed images}
    \end{subfigure}
    
    \caption{(a) The inferred object graph and (b) selected imputations of the COIL-20 dataset.
    The first column is observed images and other columns are reconstructions across missing angles.}
    \label{fig:coil_graphs}
\end{figure}

Fig.~\ref{fig:coil_graphs} shows MWGL-Missing learns meaningful graphs and imputations despite structural missingness.
For the object graph, it learns strong connections between the most similar object pairs, such as the car models, and groups other similar objects together.
The joint imputation, based on alternating Tikhonov filtering, also reasonably reconstruct the missing images by smoothing the inferred neighboring objects and views.
As we can see, imputation of symmetric objects (e.g., last row) rely on the view graph and for imputation of less symmetric objects (e.g., forth row) the object graph plays a more important role.
The limitation is that imputing a distinct object (e.g., third row) is generally challenging as it lacks meaningful neighbors.

\vspace{0.5pt}

\section{Conclusions}

In conclusion, we study the problem of Cartesian product graph learning from multi-way signals.
We establish the high-dimensional consistency guarantee for the penalized MLE, which improves the convergence rate on previous general graph Laplacian learning results.
We propose an efficient algorithm, MWGL, to solve the penalized MLE, leveraging the Cartesian product structure of the Laplacian.
Compared with several GSP and GM baselines, we demonstrate the superiority of our algorithm on both synthetic and real-world datasets. 
We also provide a joint graph learning and imputation algorithm, MWGL-Missing, and show its efficacy in the presence of structural missingness.

\putbib


\newpage
\setcounter{section}{0}

%

%

\onecolumn
\thispagestyle{plain}
\aistatstitle{Supplementary Materials}

\section{Proof of Main Theorems and Lemmas}

\subsection{Proof of Lemma 1: Efficient Computation}
We first state the following lemma which characterizes the spectral structure of the Cartesian product Laplacian.
\begin{lemma}[Eigen-decomposition of Cartesian Product]
    \label{lm:cart_eig}
    With proper ordering, the eigenvectors of the product graph Laplacian is the Kronecker product of the eigenvectors of the factor graph Laplacian
    \[
    \mathbf{U} = \mathbf{U}_1 \otimes \mathbf{U}_2,
    \] 
    and the eigenvalues of the product graph Laplacian are the Kronecker sum of the eigenvalues of the factor graph Laplacian
    \[
    \mathbf{\Lambda} = \mathbf{\Lambda}_1 \oplus \mathbf{\Lambda}_2.
    \]
\end{lemma}
Lemma.~\ref{lm:cart_eig} follows from the properties of Kronecker product \citep{barik2018spectra}.
Now we proceed to prove Lemma.~\ref{lm:efficient_h1h2}.
\begin{proof}
    We now derive the efficient computation of $\mathbf{H}_1$ that avoids the expensive large matrix inversion.
    Let the eigen-decomposition of the factor Laplacians be $\mathcal{L}\mathbf{w}_1=\mathbf{U}_1\boldsymbol\Lambda_1\mathbf{U}_1^T$ and $\mathcal{L}\mathbf{w}_2=\mathbf{U}_2\boldsymbol\Lambda_2\mathbf{U}_2^T$.
    By Lemma~\ref{lm:cart_eig}, we have the eigendecomposition of the product Laplacian 
    \begin{align}
        \mathbf{L} = \mathcal{L}\mathbf{w}_1\oplus\mathcal{L}\mathbf{w}_2 = {(\mathbf{U}_1 \otimes \mathbf{U}_2)} (\mathbf{\Lambda}_1 \oplus \mathbf{\Lambda}_2) {(\mathbf{U}_1 \otimes \mathbf{U}_2)}^T
    \end{align}
    Additionally, we notice that the eigenvectors of $\mathbf{L}^\dagger$ are also $\mathbf{U}_1\otimes\mathbf{U}_2$.
    This helps us derive the following
\begin{align}
    \mathbf{H}_1 & = \sum_{l=1}^{p_2} {(\mathbf{I}_{p_1}\otimes \mathbf{e}_{l})}^T {(\mathcal{L}\mathbf{w}_1\oplus\mathcal{L}\mathbf{w}_2)}^\dagger (\mathbf{I}_{p_1}\otimes \mathbf{e}_{l})\\
    & = \sum_{l=1}^{p_2} {(\mathbf{I}_{p_1}\otimes \mathbf{e}_{l})}^T {(\mathcal{L}\mathbf{w}_1\oplus\mathcal{L}\mathbf{w}_2)}^\dagger (\mathbf{I}_{p_1}\otimes \mathbf{e}_{l})\\
    & = \sum_{l=1}^{p_2} {(\mathbf{I}_{p_1}\otimes \mathbf{e}_{l})}^T (\mathbf{U}_1\otimes\mathbf{U}_2) {(\mathbf{\Lambda}_1\oplus\mathbf{\Lambda}_2)}^\dagger {(\mathbf{U}_1\otimes\mathbf{U}_2)}^T (\mathbf{I}_{p_1}\otimes \mathbf{e}_{l})\\
    & = \sum_{l=1}^{p_2} {(\mathbf{I}_{p_1}\otimes \mathbf{e}_{l})}^T \sum_{i=1}^{p_1}\sum_{j=1}^{p_2} \frac{1}{{[\mathbf{\Lambda_1}]}_{i,i}+{[\mathbf{\Lambda_2}]}_{j,j}}({[\mathbf{U}_1]}_{:,i}\otimes{[\mathbf{U}_2]}_{:,j}) {({[\mathbf{U}_1]}_{:,i}\otimes{[\mathbf{U}_2]}_{:,j})}^T (\mathbf{I}_{p_1}\otimes \mathbf{e}_{l})\\
    & = \sum_{l=1}^{p_2}\sum_{i=1}^{p_1}\sum_{j=1}^{p_2} {[\mathbf{U}_2]}_{i,j}^2 \frac{1}{{[\mathbf{\Lambda_1}]}_{i,i}+{[\mathbf{\Lambda_2}]}_{j,j}} {[\mathbf{U}_1]}_{:,i} {[\mathbf{U}_1]}_{:,i}^T\\
    & = \sum_{i=1}^{p_1}\sum_{j=1}^{p_2} \frac{1}{{[\mathbf{\Lambda}_1]}_{i,i}+{[\mathbf{\Lambda}_2]}_{j,j}} {[\mathbf{U}_1]}_{:,i} {[\mathbf{U}_1]}_{:,i}^T\\
    & = \sum_{j=1}^{p_2} \mathbf{U}_1 {(\mathbf{\Lambda}_1+{[\mathbf{\Lambda_2}]}_{j,j}\mathbf{I}_{p_1})}^\dagger \mathbf{U}_1^T.
\end{align}
Computation of $\mathbf{H}_2$ is derived similarly.
\end{proof}

\subsection{Proof of Theorem 2: Existence}

\begin{proof}
    Given $\mathbf{L}=\mathbf{L}_1\oplus\mathbf{L}_2$, we now prove that the global minimizer of the following penalized MLE
    \begin{equation}
    \label{eq:existence}
        \min_{\mathbf{L} \in \Omega_\mathbf{L}}  \left\{ \Tr(\mathbf{L} \mathbf{S}) - \log{\det(\mathbf{L})} + \alpha {\|\mathbf{L}\|}_{1,\mathrm{off}} \right\},
    \end{equation}
    uniquely exists.
    Provided that both the product and factor graphs are connected, The feasible set over $\mathbf{w}_1$ and $\mathbf{w}_2$ is defined as
    \begin{equation}
        \Omega_{\mathbf{w}_1,\mathbf{w}_2} := \{(\mathbf{w}_1,\mathbf{w}_2)|\mathbf{w}_1\geq0, \mathcal{L}\mathbf{w}_1+\mathbf{J}_{p_1}\in\mathcal{S}_{++}^{p_2}, \mathbf{w}_2\geq0, \mathcal{L}\mathbf{w}_2+\mathbf{J}_{p_2}\in\mathcal{S}_{++}^{p_2}\},
    \end{equation}
    where $\mathbf{J}_p = \frac{1}{p}\mathbf{1}_p\mathbf{1}_p^T$ and we have ${\logdet}^\dagger(\mathbf{L})=\logdet(\mathbf{L}+\mathbf{J}_p)$.
    The conditions $\mathcal{L}\mathbf{w}_1+\mathbf{J}_{p_1}\in\mathcal{S}_{++}^{p_1}$ and $\mathcal{L}\mathbf{w}_2+\mathbf{J}_{p_2}\in\mathcal{S}_{++}^{p_2}$ constrain that $G_1$ and $G_2$ are connected.
    Let $\{0=\lambda_1\leq\lambda_2\leq\dots\leq\lambda_p\}$ be the eigenvalues of $\mathbf{L}$ and $\mathbf{S}_1$ and $\mathbf{S}_2$ as defined in Section (3.3).
    We first consider the MLE ($\alpha=0$) and bound the negative log-likelihood $Q(\mathbf{w}_1,\mathbf{w}_2)$ as below
    {\allowdisplaybreaks
    \begin{align}
        & \Tr(\mathbf{LS})-\logdet(\mathbf{L}) \\
        = & \Tr(\mathbf{LS})-\log(\prod_{k=2}^{p}\lambda_k)\\
        \geq & \Tr(\mathbf{LS})-(p-1)\log(\sum_{k=1}^{p}\lambda_k) + (p-1)\log(p-1) \label{ieq:amgm}\\
        = & \Tr(\mathbf{L}_1\mathbf{S}_1) + \Tr(\mathbf{L}_2\mathbf{S}_2)-(p-1)\log(p_2\sum_{i=1}^{p_1}\lambda_i+p_1\sum_{j=1}^{p_2}\lambda_j) + (p-1)\log(p-1) \label{ieq:ct_prod}\\
        = & \langle \mathcal{L}^*(\mathbf{S}_1), \mathbf{w}_1 \rangle + \langle \mathcal{L}^*(\mathbf{S}_2), \mathbf{w}_2 \rangle -(p-1)\log(p_2{\|\mathbf{w}_1\|}_1+p_1{\|\mathbf{w}_2\|}_1) + (p-1)\log(\frac{p-1}{2}) \label{ieq:trace}\\
        \geq & \min (\mathcal{L}^*\Tilde{\mathbf{S}}_1 \cup \mathcal{L}^*\Tilde{\mathbf{S}}_2) (p_2{\|\mathbf{w}_1\|}_1+p_1{\|\mathbf{w}_2\|}_1) -(p-1)\log(p_2{\|\mathbf{w}_1\|}_1+p_1{\|\mathbf{w}_2\|}_1) + (p-1)\log(\frac{p-1}{2}),
    \end{align}
    }
    where ${[\Tilde{\mathbf{S}}_1]}_{i,j}=\frac{1}{p_2}{[\mathbf{S}_1]}_{i,j}$ and ${[\Tilde{\mathbf{S}}_2]}_{i,j}=\frac{1}{p_1}{[\mathbf{S}_2]}_{i,j}$.
    Inequality~\eqref{ieq:amgm} holds from the AM-GM inequality, which states that the arithmetic mean of a list of real non-negative numbers is no less than their geometric mean.
    \eqref{ieq:ct_prod} is attributed to the properties of Cartesian product graphs, and \eqref{ieq:trace} to that the summation of eigenvalues is equal to the trace of the Laplacian.
    Define the function
    \begin{equation}
    \label{eq:mle_bd}
        q(t) = \min (\mathcal{L}^*\Tilde{\mathbf{S}}_1 \cup \mathcal{L}^*\Tilde{\mathbf{S}}_2) t - (p-1)\log(t) + (p-1)\log(\frac{p-1}{2}).
    \end{equation}
    This function is lower-bounded at $t=\frac{p-1}{\min (\mathcal{L}^*\Tilde{\mathbf{S}}_1 \cup \mathcal{L}^*\Tilde{\mathbf{S}}_2)}$, so long as $\min (\mathcal{L}^*\Tilde{\mathbf{S}}_1 \cup \mathcal{L}^*\Tilde{\mathbf{S}}_2)>0$.
    Therefore, we have that the negative log-likelihood is also lower-bounded
    \begin{equation}
        Q(\mathbf{w}_1,\mathbf{w}_2) \geq h(p_2{\|\mathbf{w}_1\|}_1+p_1{\|\mathbf{w}_2\|}_1) \geq (p-1)(1+\log(\frac{\min (\mathcal{L}^*\Tilde{\mathbf{S}}_1 \cup \mathcal{L}^*\Tilde{\mathbf{S}}_2)}{2})).
    \end{equation}
    We then notice that $q(t) \rightarrow \infty$ when $t \rightarrow \infty$.
    This is followed with $Q(\mathbf{w}_1,\mathbf{w}_2)$ being coercive, since ${\|[\mathbf{w}_1,\mathbf{w}_2]\|}_2 \rightarrow \infty \leadsto p_2{\|\mathbf{w}_1\|}_1+p_1{\|\mathbf{w}_2\|}_1 \rightarrow \infty$.
    This indicates that the global minimizer exists in $\mathrm{cl}(\Omega_{\mathbf{w}_1,\mathbf{w}_2})$.
    
    Furthermore, since the open boundaries $\mathrm{cl}(\Omega_{\mathbf{w}_1,\mathbf{w}_2})\setminus \Omega_{\mathbf{w}_1,\mathbf{w}_2}$ are results of the connectivity constraint $\mathcal{L}\mathbf{w}_1+\mathbf{J}_{p_1}\succ \mathbf{O}$ and $\mathcal{L}\mathbf{w}_2+\mathbf{J}_{p_2}\succ \mathbf{O}$, we have that $\mathrm{cl}(\Omega_{\mathbf{w}_1,\mathbf{w}_2})\setminus \Omega_{\mathbf{w}_1,\mathbf{w}_2}$ is a subset of disconnected $\mathbf{w}_1$ and $\mathbf{w}_2$.
    The set of disconnected $\mathbf{w}_1$ and $\mathbf{w}_2$ is written as
    \begin{equation}
        \{(\mathbf{w}_1,\mathbf{w}_2)| \det(\mathcal{L}\mathbf{w}_1+\mathbf{J}_{p_1})=0 \vee \det(\mathcal{L}\mathbf{w}_2+\mathbf{J}_{p_2})=0 \}.
    \end{equation}
    Since for the Cartesian product, any factor graph being disconnected leads to the product graph being disconnected,
    $\forall (\mathbf{w}_1,\mathbf{w}_2)\in \mathrm{cl}(\Omega_{\mathbf{w}_1,\mathbf{w}_2})\setminus \Omega_{\mathbf{w}_1,\mathbf{w}_2}$ we have $\logdet(\mathbf{L})=-\infty \leadsto Q(\mathbf{w}_1,\mathbf{w}_2)\rightarrow\infty$. 
    This shows that any global minimizer over $\mathrm{cl}(\Omega_{\mathbf{w}_1,\mathbf{w}_2})$ do not lie on those open boundaries, therefore \eqref{eq:existence} has at least a global minimizer in $\Omega_{\mathbf{w}_1,\mathbf{w}_2}$ so long as $\min (\mathcal{L}^*\Tilde{\mathbf{S}}_1 \cup \mathcal{L}^*\Tilde{\mathbf{S}}_2)>0$.    
    $\min (\mathcal{L}^*\Tilde{\mathbf{S}}_1 \cup \mathcal{L}^*\Tilde{\mathbf{S}}_2)>0$ holds with probability 1.

    For the penalized MLE where $\alpha>0$, we slightly modify \eqref{eq:mle_bd} to obtain a new lower bound
    \begin{equation}
        q(t) = (\min (\mathcal{L}^*\Tilde{\mathbf{S}}_1 \cup \mathcal{L}^*\Tilde{\mathbf{S}}_2)+\alpha) t - (p-1)\log(t) + (p-1)\log(\frac{p-1}{2}).
    \end{equation}
    As $\min (\mathcal{L}^*\Tilde{\mathbf{S}}_1 \cup \mathcal{L}^*\Tilde{\mathbf{S}}_2)+\alpha>0$ always hold, the penalized MLE always exists.
\end{proof}

\subsection{Proof of Theorem 3: Uniqueness}
\begin{proof}
    First we show that $\Omega_{\mathbf{w}_1,\mathbf{w}_2}$ is a convex set.
    Define the feasible set of $\mathbf{w}_1$ and $\mathbf{w}_2$ as
    \begin{align}
        & \Omega_{\mathbf{w}_1}:=\{\mathbf{w}_1|\mathbf{w_1}>\mathbf{0},\mathcal{L}\mathbf{w}_1+\mathbf{J}_{p_1}\in\mathcal{S}_{++}^{p_1}\}\\
        & \Omega_{\mathbf{w}_2}:=\{\mathbf{w}_2|\mathbf{w_2}>\mathbf{0},\mathcal{L}\mathbf{w}_2+\mathbf{J}_{p_2}\in\mathcal{S}_{++}^{p_2}\}.
    \end{align}
    We can write $\Omega_{\mathbf{w}_1,\mathbf{w}_2}=\Omega_{\mathbf{w}_1}\times\Omega_{\mathbf{w}_2}$.
    Notice that both $\Omega_{\mathbf{w}_1}$ and $\Omega_{\mathbf{w}_2}$ are convex sets.
    For any $\mathbf{w}_1^{(0)},\mathbf{w}_1^{(1)}\in\Omega_{\mathbf{w}_1}$ and $\mathbf{w}_2^{(0)},\mathbf{w}_2^{(1)}\in\Omega_{\mathbf{w}_2}$
    \begin{align}
        & \mathcal{L}\mathbf{w}_1^{(a)}+\mathbf{J}_{p_1} = a (\mathcal{L}\mathbf{w}_1^{(0)}+\mathbf{J}_{p_1}) + (1-a) (\mathcal{L}\mathbf{w}_1^{(1)}+\mathbf{J}_{p_1}) \in \mathcal{S}_{++}^{p_1}, \forall 0<a<1\\
        & \mathcal{L}\mathbf{w}_2^{(b)}+\mathbf{J}_{p_2} = b (\mathcal{L}\mathbf{w}_2^{(0)}+\mathbf{J}_{p_2}) + (1-b) (\mathcal{L}\mathbf{w}_2^{(1)}+\mathbf{J}_{p_2}) \in \mathcal{S}_{++}^{p_2}, \forall 0<b<1,
    \end{align}
    where $\mathbf{w}_1^{(a)}=a\mathbf{w}_1^{(0)}+(1-a)\mathbf{w}_1^{(1)}>\mathbf{0}$ and $\mathbf{w}_2^{(b)}=b\mathbf{w}_2^{(0)}+(1-b)\mathbf{w}_2^{(1)}>\mathbf{0}$, since the PD matrices form a convex cone.
    Or one can simply realize that the linear interpolation of any two connected graphs (of the same node set) is also connected.
    Since the direct product of convex sets is a convex set, we know that $\Omega_{\mathbf{w}_1,\mathbf{w}_2}$ is a convex set.

    Then, it remains to show that $Q(\mathbf{w}_1,\mathbf{w}_2)$ is a convex function.
    Define
    \begin{equation}
        \Omega_\mathbf{w} := \{\mathbf{w}|\mathcal{L}\mathbf{w}\in\Omega_\mathbf{L} \}.
    \end{equation}
    Since there are bijections between $\Omega_\mathbf{L}$, $\Omega_\mathbf{w}$, and $\Omega_{\mathbf{w}_1,\mathbf{w}_2}$, from now on, we slightly abuse the notation of the objective function $Q$ and switch back-and-forth upon a suitable parameterization $Q(\mathbf{L})$, $Q(\mathbf{w})$, or $Q(\mathbf{w}_1,\mathbf{w}_2)$.
    Now we know that $Q(\mathbf{w})$ is a strictly convex function of $\mathbf{w}$, and $\mathbf{w}$ is an affine function of $(\mathbf{w}_1,\mathbf{w}_2)$.
    This implies that $Q(\mathbf{w}_1,\mathbf{w}_2)$ is also a strictly convex function.
    Therefore, the global minimizer of $Q$ is unique.
\end{proof}

\subsection{Proof of Theorem 4: Consistency}

\begin{proof}
    Now we prove that the penalized MLE in \eqref{eq:existence} is asymptotically consistent.
    We use a different proof from \citep{ying2021minimax} in spirit that better aligns with the popular route in literature \citep{rothman2008sparse,greenewald2019tensor}.
    Let $\mathbf{L}^*$ be the Laplacian of the true Cartesian product graph to estimate and $\mathcal{L}\mathbf{w}^*=\mathbf{L}^*$.
    Let $\mathbf{L}_1^*$ and $\mathbf{L}_2^*$ be the true factor Laplacian, where $\mathcal{L}\mathbf{w}_1^*=\mathbf{L}_1^*$, $\mathcal{L}\mathbf{w}_2^*=\mathbf{L}_2^*$, and $\mathbf{L}^*=\mathbf{L}_1^*\oplus\mathbf{L}_2^*$.
    Correspondingly, we denote the minimizer of \eqref{eq:existence} as $\hat{\mathbf{L}}=\hat{\mathbf{L}}_1\oplus\hat{\mathbf{L}}_2$, where $\hat{\mathbf{L}}=\mathcal{L}^*\hat{\mathbf{w}}$, $\hat{\mathbf{L}}_1=\mathcal{L}^*\hat{\mathbf{w}}_1$, and $\hat{\mathbf{L}}_2=\mathcal{L}^*\hat{\mathbf{w}}_2$.
    We begin with defining a set of perturbations around $\mathbf{L}^*$
    \begin{equation}
    \label{eq:delta_set}
        \mathcal{T} = \{ \Delta_\mathbf{L}| \Delta_\mathbf{L} \in \mathcal{K}_{\mathbf{L}^*}, {\| \Delta_\mathbf{L} \|}_F = cr_{n,\mathbf{p}} \},
    \end{equation}
    where $r_{n,\mathbf{p}} = \sqrt{\frac{s\log p}{n\min(p_1,p_2)}}$ for $\mathbf{p}=(p,p_1,p_2)$ and 
    \begin{equation}
        \mathcal{K}_{\mathbf{L}^*} := \{\Delta_\mathbf{L}|\mathbf{L}^*+\Delta_\mathbf{L} \in \Omega_\mathbf{L} \}.
    \end{equation}
    Define the following convex function over $\mathcal{T}$
    \begin{equation}
        F(\Delta_\mathbf{L}) = Q(\mathbf{L}^*+\Delta_\mathbf{L}) - Q(\mathbf{L}^*).
    \end{equation}
    Our goal now is to show that 
    \begin{equation}
    \label{ieq:main_consistent}
        F(\Delta_\mathbf{L}) > 0, \forall \Delta_\mathbf{L} \in \mathcal{T}.
    \end{equation}

    To see the rationale behind \eqref{ieq:main_consistent}, notice that
    \begin{equation}
        F(\hat{\mathbf{L}}-\mathbf{L}^*) = Q(\hat{\mathbf{L}})-Q(\mathbf{L}^*) \leq 0,
    \end{equation}
    since $\hat{\mathbf{L}}$ minimize $Q(\mathbf{L})$.
    Provided that $F(\Delta_\mathbf{L})$ is a convex function, \eqref{ieq:main_consistent} ultimately implies that 
    \begin{equation}
    \label{ieq:ult_consistent}
        {\|\hat{\mathbf{L}}-\mathbf{L}^*\|}_F \leq cr_{n,\mathbf{p}}.
    \end{equation}

    To prove this is true, we first prove the following
    \begin{equation}
        F(\Delta_\mathbf{L})>0,\forall \Delta_\mathbf{L}\in\mathcal{K}_{\mathbf{L}^*}, {\|\Delta_\mathbf{L}\|}_F>cr_{n,\mathbf{p}}.
    \end{equation}
    By contradiction, suppose there exists a $\Delta_\mathbf{L}'\in\mathcal{K}_{\mathbf{L}^*}$, such that ${\|\Delta_\mathbf{L}'\|}_F>cr_{n,\mathbf{p}}$ and $F(\Delta_\mathbf{L}')<0$.
    Let $\theta=\frac{cr_{n,\mathbf{p}}}{{\|\Delta_\mathbf{L}'\|}_F}<1$.
    Then
    \begin{equation}
        F(\theta\Delta_\mathbf{L}') = F((1-\theta)\mathbf{O}+\theta\Delta_\mathbf{L}') \leq (1-\theta)F(\mathbf{O}) + \theta F(\Delta_\mathbf{L}') = \theta F(\Delta_\mathbf{L}') < 0.
    \end{equation}
    This contradicts with \eqref{ieq:main_consistent} since $\theta\Delta_\mathbf{L}'\in\mathcal{T}$.
    Thus \eqref{ieq:ult_consistent} must holds under \eqref{ieq:main_consistent}.

    Now we move forward to prove \eqref{ieq:main_consistent}.
    We write out $F(\Delta_\mathbf{L})$
    \begin{equation}
        F(\Delta_\mathbf{L}) = \Tr(\Delta_\mathbf{L}\mathbf{S}) - (\logdet(\mathbf{L}^*+\Delta_\mathbf{L}+\mathbf{J}_p) - \logdet(\mathbf{L}^*+\mathbf{J}_p)) + \alpha ({\|\mathbf{L}^*+\Delta_\mathbf{L}\|}_{1,\mathrm{off}}-{\|\mathbf{L}^*\|}_{1,\mathrm{off}}).
    \end{equation}
    Consider the Taylor's expansion of $\logdet(\mathbf{L}^*+\nu\Delta_\mathbf{L}+\mathbf{J}_p)$ with the integral remainder
    \begin{equation}
        \logdet(\mathbf{L}^*+\Delta_\mathbf{L}+\mathbf{J}_p) - \logdet(\mathbf{L}^*+\mathbf{J}_p) = \Tr({(\mathbf{L}^*+\mathbf{J}_p)}^{-1}\Delta_\mathbf{L}) + \int_0^1 (1-\nu) \nabla_\nu^2 \logdet(\mathbf{L}^*+\nu\Delta_\mathbf{L}+\mathbf{J}_p) d\nu,
    \end{equation}
    and further the remainder 
    \begin{equation}
        \int_0^1 (1-\nu) \nabla_\nu^2 \logdet(\mathbf{L}^*+\nu\Delta_\mathbf{L}+\mathbf{J}_p) d\nu = 
        -{\mathrm{vec}(\Delta_\mathbf{L})}^T (\int_0^1 (1-\nu) {(\mathbf{L}^*+\nu\Delta_\mathbf{L}+\mathbf{J}_p)}^{-1} \otimes {(\mathbf{L}^*+\nu\Delta_\mathbf{L}+\mathbf{J}_p)}^{-1} d\nu) \mathrm{vec}(\Delta_\mathbf{L}).
    \end{equation}
    Therefore we have
    \begin{align}
        F(\Delta_\mathbf{L}) = I_1 + I_2 + I_3,
    \end{align}
    where
    \begin{align}
        I_1 & = \Tr(\Delta_\mathbf{L}(\mathbf{S}-{(\mathbf{L}^*+\mathbf{J}_p)}^{-1})),\\
        I_2 & = {\mathrm{vec}(\Delta_\mathbf{L})}^T (\int_0^1 (1-\nu) {(\mathbf{L}^*+\nu\Delta_\mathbf{L}+\mathbf{J}_p)}^{-1} \otimes {(\mathbf{L}^*+\nu\Delta_\mathbf{L}+\mathbf{J}_p)}^{-1} d\nu) \mathrm{vec}(\Delta_\mathbf{L}),\\
        I_3 & = \alpha ({\|\mathbf{L}^*+\Delta_\mathbf{L}\|}_{1,\mathrm{off}}-{\|\mathbf{L}^*\|}_{1,\mathrm{off}}).
    \end{align}

    We now bound each term separately. 

    \paragraph{Bound $I_1$:}
    We follow the argument in \cite{ying2020nonconvex} and assume that the graph signals are sampled from the process referred to as conditioning by Kriging \citep{rue2005gaussian}.
    This process first sample from the proper GMRF $\bm{\mathit{x}}\sim\mathcal{N}(\mathbf{0},{(\mathbf{L}^*+\mathbf{J}_p)}^{-1})$, then correct these samples by subtracting their mean to make them DC-intrinsic
    \begin{equation}
        \Bar{\mathbf{x}} = \mathbf{x} -\frac{1}{p}\mathbf{1}\mathbf{1}^T\mathbf{x}.
    \end{equation}
    Let $\mathbf{\Sigma}={(\mathbf{L}^*+\mathbf{J}_p)}^{-1}$ be the covariance matrix of the original proper GMRF. 
    Since $\Delta_\mathbf{L} \in \mathcal{K}_{\mathbf{L}^*}$, we have
    \begin{align}
        I_1 & = \Tr((\Delta_{\mathbf{L}_1} \oplus \Delta_{\mathbf{L}_2}) (\mathbf{S}-{(\mathbf{L}^*+\mathbf{J}_p)}^{-1}))\\
        & = \Tr(\Delta_{\mathbf{L}_1} (\mathbf{S}_1-\mathbf{\Sigma}_1)) + \Tr(\Delta_{\mathbf{L}_2} (\mathbf{S}_2-\mathbf{\Sigma}_2))\\
        & = \Delta_{\mathbf{w}_1}^T \mathcal{L}^*(\mathbf{S}_1-\mathbf{\Sigma}_1) + \Delta_{\mathbf{w}_2}^T \mathcal{L}^*(\mathbf{S}_2-\mathbf{\Sigma}_2)\\
        & = p_2 \Delta_{\mathbf{w}_1}^T \mathcal{L}^*(\Tilde{\mathbf{S}}_1-\Tilde{\mathbf{\Sigma}}_1) + p_1 \Delta_{\mathbf{w}_2}^T \mathcal{L}^*(\Tilde{\mathbf{S}}_2-\Tilde{\mathbf{\Sigma}}_2).
    \end{align}
    where $\Delta_{\mathbf{L}}=\Delta_{\mathbf{L}_1}\oplus\Delta_{\mathbf{L}_2}$ and ${[\Tilde{\mathbf{\Sigma}}_1]}_{i,j}=\frac{1}{p_2}{[\mathbf{\Sigma}_1]}_{i,j}$ and ${[\Tilde{\mathbf{\Sigma}}_2]}_{i,j}=\frac{1}{p_1}{[\mathbf{\Sigma}_2]}_{i,j}$.
    $\mathbf{\Sigma}_1$ and $\mathbf{\Sigma}_2$ are defined similarly as in the Section~\ref{sec:pgd}.
    \begin{align}
        \mathbf{\Sigma}_1 = \sum_{l=1}^{p_2} {(\mathbf{I}_{p_1}\otimes \mathbf{e}_{p_2}^l)}^T \mathbf{\Sigma} (\mathbf{I}_{p_1}\otimes \mathbf{e}_{p_2}^l), \\
        \mathbf{\Sigma}_2 = \sum_{l=1}^{p_1} {(\mathbf{e}_{p_1}^l\otimes \mathbf{I}_{p_2})}^T \mathbf{\Sigma} (\mathbf{e}_{p_1}^l\otimes \mathbf{I}_{p_2}).
    \end{align}
    We then have
    \begin{align}
        & {[\mathcal{L}^*\mathbf{S}_1]}_{i-j+\frac{1}{2}(j-1)(2p_1-j)} = \frac{1}{n}\sum_{k=1}^n\sum_{l=1}^{p_2} {({[\mathbf{x}_k]}_{(i-1)p_2+l}-{[\mathbf{x}_k]}_{(j-1)p_2+l})}^2,\forall 1\leq j<i\leq p_1,\\
        & {[\mathcal{L}^*\mathbf{S}_2]}_{i-j+\frac{1}{2}(j-1)(2p_2-j)} = \frac{1}{n}\sum_{k=1}^n\sum_{l=1}^{p_1} {({[\mathbf{x}_k]}_{(l-1)p_2+i}-{[\mathbf{x}_k]}_{(l-1)p_2+j})}^2,\forall 1\leq j<i\leq p_2.
    \end{align}
    
    Here we focus on $\mathcal{L}^*\mathbf{S}_1$, and results on $\mathcal{L}^*\mathbf{S}_2$ can be derived similarly.
    Let $m_1=i-j+\frac{1}{2}(j-1)(2p_1-j)$.
    We rewrite ${[\mathcal{L}^*\Tilde{\mathbf{S}}_1]}_{m_1}$ into the quadratic form of the entries of $\mathbf{x}$
    \begin{equation}
        {[\mathcal{L}^*\mathbf{S}_1]}_{m_1} = \frac{1}{n} \sum_{k=1}^n \mathbf{x}_k^T (\mathcal{L} \mathbf{e}_{m}\otimes\mathbf{I}_{p_2}) \mathbf{x}_k,\forall 1\leq j<i\leq p_1,
    \end{equation}
    where ${\mathbf{e}_{m_1}}\in\mathbb{R}^{\frac{p_1(p_1-1)}{2}}$ has one in the $\{i-j+\frac{1}{2}(j-1)(2p_1-j)\}$-th entry and zeros otherwise.
    Let $\mathbf{x}_k=\mathbf{\Sigma}^{\frac{1}{2}}\mathbf{z}_k$, where $\bm{\mathit{z}_k}\sim\mathcal{N}(\mathbf{0},\mathbf{I}_p)$ is the source signal of the GSP system.
    We then write the above quadratic as
    \begin{equation}
    \label{eq:ls_quad}
        {[\mathcal{L}^*\mathbf{S}_1]}_{m_1} = \frac{1}{n} \sum_{k=1}^n {\mathbf{z}}_k^T \mathbf{\Sigma}^{\frac{1}{2}} (\mathcal{L} \mathbf{e}_{m_1}\otimes\mathbf{I}_{p_2}) \mathbf{\Sigma}^{\frac{1}{2}} \mathbf{z}_k,\forall 1\leq j<i\leq p_1,
    \end{equation}
    Let $\mathbf{M}_{i,j} = \mathbf{\Sigma}^{\frac{1}{2}} (\mathcal{L} \mathbf{e}_{m_1}\otimes\mathbf{I}_{p_2})\mathbf{\Sigma}^{\frac{1}{2}}$.
    By the Hanson-Wright inequality \cite{hanson1971bound,rudelson2013hanson}, we have
    \begin{align}
        \mathbb{P} \left\{ |\frac{1}{n}\sum_{k=1}^n \mathbf{z}_k^T \mathbf{M}_{i,j} \mathbf{z}_k-\mathbb{E}[\frac{1}{n}\sum_{k=1}^n \mathbf{z}_k^T \mathbf{M}_{i,j} \mathbf{z}_k]|>h \right\}  & \leq 2\exp{\left[-c_1 \min{(\frac{nh^2}{K^4{\|\mathbf{M}_{i,j}\|}_F^2},\frac{nh}{K^2{\|\mathbf{M}_{i,j}\|}_2})}\right]} \label{ieq:hanson_wright}\\
        & \leq 2\exp{\left[-c_1 \min{(\frac{nh^2}{4K^4p_2{\|\mathbf{\Sigma}\|}_2^2},\frac{nh}{2K^2{\|\mathbf{\Sigma}\|}_2})}\right]} \label{ieq:spectral_norm}\\
        & \leq 2\exp{\left[-c_1 \min{(\frac{nh^2}{64p_2{\|\mathbf{\Sigma}\|}_2^2},\frac{nh}{8{\|\mathbf{\Sigma}\|}_2})}\right]}, \label{eq:sub_gaussian}
    \end{align}
    where $K=2$ is the sub-Gaussian norm of $\mathbf{z}_k$.
    \eqref{ieq:spectral_norm} holds by the properties of matrix norms and the trace inequalities \citep{fang1994inequalities}
    \begin{align}
        & {\|\mathbf{M}_{i,j}\|}_F^2 \leq {\|\mathbf{\Sigma}^{\frac{1}{2}}\|}_2^2 {\|\mathcal{L}\mathbf{e}_{m_1}\otimes\mathbf{I}_{p_2})\|}_F^2 {\|\mathbf{\Sigma}^{\frac{1}{2}}\|}_2^2 = 4p_2{\|\mathbf{\Sigma}\|}_2^2, \\
        & {\|\mathbf{M}_{i,j}\|}_2 \leq {\|\mathbf{\Sigma}^{\frac{1}{2}}\|}_2 {\|\mathcal{L}\mathbf{e}_{m_1}\otimes\mathbf{I}_{p_2})\|}_2 {\|\mathbf{\Sigma}^{\frac{1}{2}}\|}_2 = 2{\|\mathbf{\Sigma}\|}_2 ,
    \end{align}
    where ${\|\mathcal{L}\mathbf{e}_{m_1}\otimes\mathbf{I}_{p_2})\|}_F^2=4p_2$ and ${\|\mathcal{L}\mathbf{e}_{m_1}\otimes\mathbf{I}_{p_2})\|}_2={\|\mathcal{L}\mathbf{e}_{m_1}\|}_2=2$.
    Let $\epsilon=\frac{h}{\sqrt{p_2}{{\|\mathbf{\Sigma}\|}_2}}$ and plug \eqref{eq:ls_quad} into \eqref{eq:sub_gaussian}
    \begin{equation}
        \mathbb{P} \left\{ |{[\mathcal{L}^*\mathbf{S}_1]}_{m_1}-\mathbb{E}[{[\mathcal{L}^*\mathbf{S}_1]}_{m_1}]|>\epsilon \sqrt{p_2}{\|\mathbf{\Sigma}\|}_2 \right\}  \leq 2\exp{(-\frac{cn\epsilon^2}{64})}, \forall \epsilon\leq8\sqrt{p_2}.
    \end{equation}
    
    Meanwhile 
    \begin{align}
        \mathbb{E}[{[\mathcal{L}^*\mathbf{S}_1]}_{m_1}] & = \frac{1}{n} \sum_{k=1}^n\sum_{l=1}^{p_2} \mathbb{E}[{({[\mathbf{x}_k]}_{(i-1)p_2+l}-{[\mathbf{x}_k]}_{(j-1)p_2+l})}^2] \\
        & = \frac{1}{n} \sum_{k=1}^n\sum_{l=1}^{p_2} \mathbb{E}[{[\mathbf{x}_k]}_{(i-1)p_2+l}^2] - 2\mathbb{E}[{[\mathbf{x}_k]}_{(i-1)p_2+l} {[\mathbf{x}_k]}_{(j-1)p_2+l}] +\mathbb{E}[{[\mathbf{x}_k]}_{(j-1)p_2+l}^2]\\
        & = {[\mathbf{\Sigma}_1]}_{i,i}-{[\mathbf{\Sigma}_1]}_{i,j}-{[\mathbf{\Sigma}_1]}_{j,i}+{[\mathbf{\Sigma}_1]}_{j,j} \\
        & = {[\mathcal{L}^*\mathbf{\Sigma}_1]}_{m_1}.
    \end{align}
    
    Therefore 
    \begin{align}
        \mathbb{P} \left\{|{[\mathcal{L}^*(\mathbf{S}_1-\mathbf{\Sigma}_1)]}_{m_1}| > \epsilon \sqrt{p_2}{\|\mathbf{\Sigma}\|}_2\right\}  \leq 2\exp{(-\frac{c_1n\epsilon^2}{64})}, \forall \epsilon\leq8\sqrt{p_2},
    \end{align}
    and we reach the following concentration result for $\mathcal{L}^*\Tilde{\mathbf{S}}_1$
    \begin{align}
        \mathbb{P} \left\{ |{[\mathcal{L}^*(\Tilde{\mathbf{S}}_1-\Tilde{\mathbf{\Sigma}}_1)]}_{m_1}| > \frac{\epsilon {\|\mathbf{\Sigma}\|}_2}{\sqrt{p_2}} \right\}  \leq 2\exp{(-\frac{c_1n\epsilon^2}{64})}, \forall \epsilon\leq8\sqrt{p_2},
    \end{align}
    Similarly for $\mathcal{L}^*\Tilde{\mathbf{S}}_2$ we derive for $m_2=i-j+\frac{1}{2}(j-1)(2p_2-j)$
    \begin{equation}
        \mathbb{P} \left\{ |{[\mathcal{L}^*(\Tilde{\mathbf{S}}_2-\Tilde{\mathbf{\Sigma}}_2)]}_{m_2}| > \frac{\epsilon {\|\mathbf{\Sigma}\|}_2}{\sqrt{p_1}} \right\}  \leq 2\exp{(-\frac{c_1n\epsilon^2}{64})}, \forall \epsilon\leq8\sqrt{p_1}.
    \end{equation}
    
    By the union bound 
    \begin{align}
        & \mathbb{P} \left\{\max{\left[ |\mathcal{L}^*(\Tilde{\mathbf{S}}_1-\Tilde{\mathbf{\Sigma}}_1)|,|\mathcal{L}^*(\Tilde{\mathbf{S}}_2-\Tilde{\mathbf{\Sigma}}_2)| \right]} > \frac{\epsilon{\|\mathbf{\Sigma}\|}_2}{\sqrt{\min(p_1,p_2)}} \right\} \\
        \leq & \mathbb{P} \left\{\max_{m_1}| {[\mathcal{L}^*(\Tilde{\mathbf{S}}_1-\Tilde{\mathbf{\Sigma}}_1)]}_{m_1}| > \frac{\epsilon{\|\mathbf{\Sigma}\|}_2}{\sqrt{p_2}} \right\}  + \mathbb{P} \left\{ \max_{m_2} | {[\mathcal{L}^*(\Tilde{\mathbf{S}}_2-\Tilde{\mathbf{\Sigma}}_2)]}_{m_2} | > \frac{\epsilon{\|\mathbf{\Sigma}\|}_2}{\sqrt{p_1}} \right\}  \\
        \leq & \sum_{m_1=1}^{\frac{p_1(p_1-1)}{2}} 2\exp{(-\frac{c_1n\epsilon^2}{64})} + \sum_{m_2=1}^{\frac{p_2(p_2-1)}{2}} 2\exp{(-\frac{c_1n\epsilon^2}{64})} \\
        \leq & 2{\max}^2{(p_1,p_2)}\exp{(-\frac{c_1n\epsilon^2}{64})}.
    \end{align}
    By calculation we then have
    \begin{equation}
    \label{ieq:f1_prob}
        \mathbb{P} \left\{\max{\left[ |\mathcal{L}^*(\Tilde{\mathbf{S}}_1-\Tilde{\mathbf{\Sigma}}_1)|,|\mathcal{L}^*(\Tilde{\mathbf{S}}_2-\Tilde{\mathbf{\Sigma}}_2)| \right]} \leq \frac{\epsilon{\|\mathbf{\Sigma}\|}_2}{\sqrt{\min(p_1,p_2)}} \right\} \geq 1-2{\max}^2{(p_1,p_2)}\exp{(-\frac{c_1n\epsilon^2}{64})}.
    \end{equation}
    So with the probability stated in \eqref{ieq:f1_prob}, we derive the following lower bound for $I_1$
    \begin{align}
        I_1 & = p_2\Tr(\Delta_{\mathbf{L}_1} (\Tilde{\mathbf{S}}_1-\Tilde{\mathbf{\Sigma}}_1)) + p_1\Tr(\Delta_{\mathbf{L}_2} (\Tilde{\mathbf{S}}_2-\Tilde{\mathbf{\Sigma}}_2)) \\
        & \geq - p_2|\Delta_{\mathbf{w}_1}^T \mathcal{L}^*(\Tilde{\mathbf{S}}_1-\Tilde{\mathbf{\Sigma}}_1)| - p_1|\Delta_{\mathbf{w}_2}^T \mathcal{L}^*(\Tilde{\mathbf{S}}_2-\Tilde{\mathbf{\Sigma}}_2)|\\
        & \geq -\max {\left[|\mathcal{L}^*(\Tilde{\mathbf{S}}_1-\Tilde{\mathbf{\Sigma}}_1)|, |\mathcal{L}^*(\Tilde{\mathbf{S}}_2-\Tilde{\mathbf{\Sigma}}_2)|\right]}  (p_2|\Delta_{\mathbf{w}_1}|+p_1|\Delta_{\mathbf{w}_2}|)\\
        & \geq -\frac{\epsilon{\|\mathbf{\Sigma}\|}_2}{\sqrt{\min(p_1,p_2)}} |\Delta_\mathbf{w}|.
    \end{align}

    \paragraph{Bound $I_2$:}
    From the min-max theorem, we have
    \begin{equation}
        I_2 \geq {\|\Delta_\mathbf{L}\|}_F^2 \lambda_{\mathrm{min}} (\int_0^1 (1-\nu) {(\mathbf{L}^*+\nu\Delta_\mathbf{L}+\mathbf{J}_p)}^{-1} \otimes {(\mathbf{L}^*+\nu\Delta_\mathbf{L}+\mathbf{J}_p)}^{-1} d\nu).
    \end{equation}
    Then given the convexity of $\lambda_{\mathrm{max}}(\cdot)$ and concavity of $\lambda_{\mathrm{min}}(\cdot)$
    {\allowdisplaybreaks
    \begin{align}
        & \lambda_{\mathrm{min}} (\int_0^1 (1-\nu) {(\mathbf{L}^*+\nu\Delta_\mathbf{L}+\mathbf{J}_p)}^{-1} \otimes {(\mathbf{L}^*+\nu\Delta_\mathbf{L}+\mathbf{J}_p)}^{-1} d\nu) \\
        \geq & \int_0^1 (1-\nu) \lambda_{\mathrm{min}}^2{(\mathbf{L}^*+\nu\Delta_\mathbf{L}+\mathbf{J}_p)}^{-1} d\nu \\
        \geq & \min_{\nu\in [0,1]} \left[ \lambda_{\mathrm{min}}^2{(\mathbf{L}^*+\nu\Delta_\mathbf{L}+\mathbf{J}_p)}^{-1}\right] \int_0^1 (1-\nu) d\nu \\
        = & \frac{1}{2} \min_{\nu\in [0,1]} \left[ \frac{1}{\lambda_{\mathrm{max}}^2(\mathbf{L}^*+\nu\Delta_\mathbf{L}+\mathbf{J}_p)} \right] \\
        = & \frac{1}{2\max_{\nu\in [0,1]} \left[ \lambda_{\mathrm{max}}^2(\mathbf{L}^*+\nu\Delta_\mathbf{L}+\mathbf{J}_p)\right] } \\
        \geq & \frac{1}{2\max_{\nu\in [0,1]}^2 {\left[(\lambda_{\mathrm{max}}(\mathbf{L}^*+\mathbf{J}_p)+{\|\nu\Delta_\mathbf{L}\|}_2)\right]}} \\
        = & \frac{1}{2{(\lambda_{\mathrm{max}}(\mathbf{L}^*+\mathbf{J}_p)+{\|\Delta_\mathbf{L}\|}_2)}^2}
    \end{align}
    }
    Then with $n$ sufficiently large 
    \begin{equation}
        n \geq \frac{c^2s\log p}{\lambda^2_{\mathrm{max}}(\mathbf{L}^*+\mathbf{J}_p)\min(p_1,p_2)},
    \end{equation}
    such that ${\|\Delta_\mathbf{L}\|}_2 \leq {\|\Delta_\mathbf{L}\|}_F \leq \lambda_{\mathrm{max}}(\mathbf{L}^*+\mathbf{J}_p)$, we obtain a lower bound for $I_2$
    \begin{align}
        I_2 & \geq \frac{{\|\Delta_\mathbf{L}\|}_F^2}{2{(\lambda_{\mathrm{max}}(\mathbf{L}^*+\mathbf{J}_p)+{\|\Delta_\mathbf{L}\|}_2)}^2}\\
        & \geq \frac{{\|\Delta_\mathbf{L}\|}_F^2}{8 \lambda^2_{\mathrm{max}}(\mathbf{L}^*+\mathbf{J}_p)}.
    \end{align}

    \paragraph{Bound $I_3$:}
    To bound $I_3$, we use triangular inequality
    \begin{align}
        {\|\mathbf{L}^*+\Delta_\mathbf{L}\|}_{1,\mathrm{off}}-{\|\mathbf{L}^*\|}_{1,\mathrm{off}} = {\|\mathbf{L}^*+\Delta_\mathbf{L}\|}_{1,\mathcal{A}}+{\|\Delta_\mathbf{L}\|}_{1,\mathcal{A}^\complement}-{\|\mathbf{L}^*\|}_{1,\mathcal{A}}
        \geq {\|\Delta_\mathbf{L}\|}_{1,\mathcal{A}^\complement}-{\|\Delta_\mathbf{L}\|}_{1,\mathcal{A}},
    \end{align}
    to obtain
    \begin{equation}
        I_3 \geq \alpha ({\|\Delta_\mathbf{L}\|}_{1,\mathcal{A}^\complement}-{\|\Delta_\mathbf{L}\|}_{1,\mathcal{A}})
        = 2\alpha{|\Delta_\mathbf{w}|}_{1,\mathcal{A}^\complement} - 2\alpha{|\Delta_\mathbf{w}|}_{1,\mathcal{A}}.
    \end{equation}

    \paragraph{Bound $I_1+I_2+I_3$:}
    So overall
    \begin{align}
        F(\Delta_\mathbf{L}) & \geq - \frac{\epsilon{\|\mathbf{\Sigma}\|}_2}{\sqrt{\min(p_1,p_2)}} {| \Delta_\mathbf{w}|}_1 + \frac{{\|\Delta_\mathbf{L}\|}_F^2}{8 \lambda^2_{\mathrm{max}}(\mathbf{L}^*+\mathbf{J}_p)} + 2\alpha{|\Delta_\mathbf{w}|}_{1,\mathcal{A}^\complement} - 2\alpha{|\Delta_\mathbf{w}|}_{1,\mathcal{A}}\\
        & = \frac{{\|\Delta_\mathbf{L}\|}_F^2}{8 \lambda^2_{\mathrm{max}}(\mathbf{L}^*+\mathbf{J}_p)} - (\frac{\epsilon{\|\mathbf{\Sigma}\|}_2}{\sqrt{\min(p_1,p_2)}}-2\alpha){|\Delta_\mathbf{w}|}_{1,\mathcal{A}^\complement} - (\frac{\epsilon{\|\mathbf{\Sigma}\|}_2}{\sqrt{\min(p_1,p_2)}}+2\alpha){|\Delta_\mathbf{w}|}_{1,\mathcal{A}}.
    \end{align}
    Let $\epsilon=c_2\sqrt{\frac{\log p}{n}}$ with sufficiently large
    \begin{equation}
        n \geq \frac{c_2^2\log p}{64\min(p_1,p_2)},
    \end{equation}
    so that $\epsilon \leq 8\sqrt{\min(p_1,p_2)}$.
    Then choose
    \begin{equation}
        \alpha \geq \frac{c_2{\|\mathbf{\Sigma}\|}_2}{2}\sqrt{\frac{\log p}{n\min(p_1,p_2)}},
    \end{equation}
    such that $\epsilon{\|\mathbf{\Sigma}\|}_2-2\alpha\sqrt{\min(p_1,p_2)}\leq 0$.
    Also note that
    \begin{equation}
        {|\Delta_\mathbf{w}|}_{1,\mathcal{A}} \leq \sqrt{s} {\|\Delta_\mathbf{w}\|}_{2,\mathcal{A}} \leq\sqrt{s} {\|\Delta_\mathbf{w}\|}_2 \leq \sqrt{\frac{s}{2}}{\|\Delta_\mathbf{L}\|}_F,
    \end{equation}
    then with $\frac{\alpha}{\gamma} = \frac{c_2{\|\mathbf{\Sigma}\|}_2}{2}\sqrt{\frac{\log p}{n\min(p_1,p_2)}}$ we obtain
    \begin{align}
        F(\Delta_\mathbf{L}) & \geq \frac{{\|\Delta_\mathbf{L}\|}_F^2}{8 \lambda^2_{\mathrm{max}}(\mathbf{L}^*+\mathbf{J}_p)} - (1+\gamma)c_2{\|\mathbf{\Sigma}\|}_2\sqrt{\frac{\log p}{n\min(p_1,p_2)}}{|\Delta_\mathbf{w}|}_{1,\mathcal{A}}\\
        & \geq {\|\Delta_\mathbf{L}\|}_F^2(\frac{1}{8 \lambda^2_{\mathrm{max}}(\mathbf{L}^*+\mathbf{J}_p)} - (1+\gamma)c_2{\|\mathbf{\Sigma}\|}_2\sqrt{\frac{s\log p}{2n\min(p_1,p_2)}}{\|\Delta_\mathbf{L}\|}_F^{-1})\\
        & = {\|\Delta_\mathbf{L}\|}_F^2(\frac{1}{8 \lambda^2_{\mathrm{max}}(\mathbf{L}^*+\mathbf{J}_p)} - (1+\gamma)\frac{c_2{\|\mathbf{\Sigma}\|}_2}{\sqrt{2}c})\\
        & > 0,
    \end{align}
    so long as $c$ is sufficiently large
    \begin{equation}
        c > 4\sqrt{2}(1+\gamma)c_2{\|\mathbf{\Sigma}\|}_2 \lambda^2_{\mathrm{max}}(\mathbf{L}^*+\mathbf{J}_p).
    \end{equation}
    This holds with probability at least
    \begin{align}
        & \mathbb{P} \left\{\max{\left[ |\mathcal{L}^*(\Tilde{\mathbf{S}}_1-\Tilde{\mathbf{\Sigma}}_1)|,|\mathcal{L}^*(\Tilde{\mathbf{S}}_2-\Tilde{\mathbf{\Sigma}}_2)| \right]} \leq \frac{\epsilon{\|\mathbf{\Sigma}\|}_2}{\sqrt{\min(p_1,p_2)}} \right\} \\
        \geq & 1-2{\max}^2{(p_1,p_2)}\exp{(-\frac{c_1n\epsilon^2}{64})}\\
        = & 1-2\exp{\left[2\log \left[\max{(p_1,p_2)}\right]-\frac{c_1c_2^2}{64}\log p\right]}\\
        \geq &1-2\exp{(-c'\log p)},
    \end{align}
    where $c' = \frac{c_1c_2^2}{64}-2$.
    $\gamma \geq 1$ here is a tuning parameter.
    Setting $\gamma=1$ retrieves Theorem.~\ref{th:consistency}.
\end{proof}

\section{Experimental Details}

We now detail our experimental settings.
To initialize $\mathbf{w}_1$ and $\mathbf{w}_2$, we first calculate $\mathbf{S}_1^{-1}$ and $\mathbf{S}_2^{-1}$ to obtain an initial guess of the factor precision matrices.
The non-Laplacian matrices are then processed as described in Sec.~\ref{sec:exp_synthetic} to initialize $\mathbf{w}_1={(-\mathrm{tril}(\mathbf{S}_1^{-1},-1))}_+$ and $\mathbf{w}_2={(-\mathrm{tril}(\mathbf{S}_2^{-1},-1))}_+$.
When there are missing values, we use the largest sub-columns/rows of $\mathbf{X}_k$ that do not include missing entries to compute 
\begin{align}
    & \mathbf{S}_1=\frac{1}{n}\sum_{k=1}^n{[\mathbf{X}_k]}_{:,\Psi_c}{{[\mathbf{X}_k]}_{:,\Psi_c}}^T,
    & \mathbf{S}_2=\frac{1}{n}\sum_{k=1}^n{{[\mathbf{X}_k]}_{\Psi_r,:}}^T{[\mathbf{X}_k]}_{\Psi_r,:}.
\end{align}
Here $\Psi_c$ and $\Psi_r$ are defined as the sets of all the columns and rows that do not contain missing values.
For the initial imputation of a node $(r_1,r_2)\in\Psi^\complement$, we consider the set of node pairs $\Psi_{(r_1,r_2)} = \{(r_1,i_2)\not\in \Psi^\complement\} \vee \{(i_1,r_2)\not\in \Psi^\complement\}$,
and use
\begin{equation}
    {[\mathbf{X}_k]}_{r_1,r_2} = \frac{1}{|\Psi_{(r_1,r_2)}|}\sum_{(i_1,i_2)\in \Psi_{(r_1,r_2)}}{[\mathbf{X}_k]}_{i_1,i_2},
\end{equation}
i.e. average of all non-missing entries that belong to the same column or row with itself.
For all experiments, we set the learning rate $\eta$ of project gradient descent to 1e-3 and the tolerance $\epsilon$ to be 1e-6.
For pre-processing steps, we normalize the COIL-20 images with $\frac{\mathbf{X}}{255}-0.5$ and remove the station and hour means of the Molene data.


We use official implementations for the PGL, BiGLasso, and TeraLasso baselines.
For the PST baseline, since the original paper proposed to learn the eigenvectors of factor adjacency matrices \citep{einizade2023learning}, we implement an adapted algorithm that learns eigenvectors of factor Laplacian matrices.
Such adaptation is presented in the Sec. 5C of \citep{segarra2017network}.

\subsection{PR-AUC of Edge Estimation of Synthetic Graphs}

\begin{figure}[hbt!]
\centering
\includegraphics[width=\textwidth]{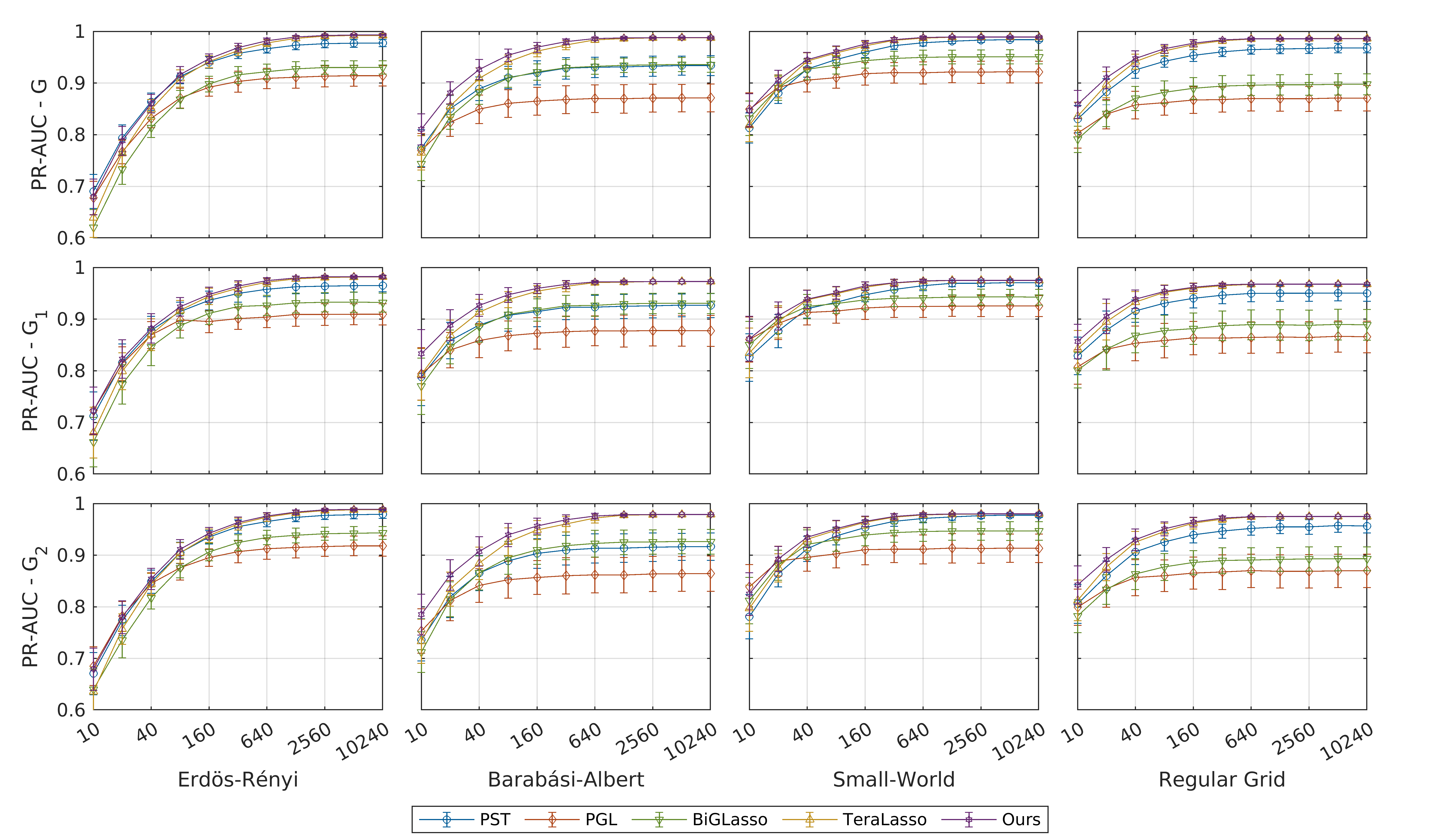}
\caption{Comparison of different methods on synthetic data in various scenarios.
Each sub-figure shows the PR-AUC of edge estimation as $n$ increases.}
\label{fig:prauc}
\end{figure}

To compute PR-AUC, we use $\mathds{1}_{\mathbf{w}>\rho}(\hat{\mathbf{w}})$ as the binary edge predictions of an increasing series of threshold $\rho$ and calculate area of the precision-recall curves.
The results are shown in Fig.~\ref{fig:prauc}.
We can see that MWGL again outperforms all the baselines on different settings.

\subsection{Synthetic Experiments with Varying Factor Size}

\begin{figure}[hbt!]
\centering
\includegraphics[width=\textwidth]{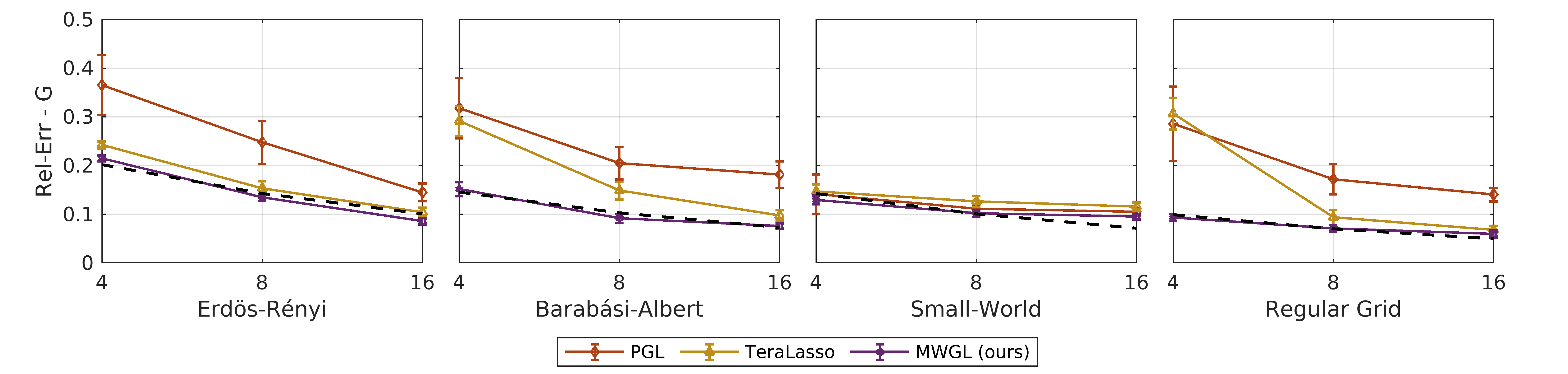}
\caption{Comparison of different methods on synthetic data in various scenarios.
Each sub-figure shows the trend of Rel-Err of the product or factor Laplacian matrices as $\min(p_1,p_2))$ increases.
Black dash lines fit the theory in \eqref{ieq:convergence_rate} to our results.}
\label{fig:fnorm_minp}
\end{figure}

We now evaluate MWGL on synthetic data with fixed $p$ but varying $p_1$ and $p_2$.
Our main goal is to verify the convergence rate in \ref{th:consistency} as a function of $\min{(p_1,p_2)}$, but we also compare MWGL with PGL and TeraLasso. 
We fix the size of product graphs to be $p=256$ and set $p_1$ to be 4, 8, or 16, and $p_2$ to be 64, 32, and 16, respectively.
We use the same graph models stated in Sec.~\ref{sec:exp_synthetic} to generate factor graphs.
For regular grids, we always set the width to 2 and the height correspondingly.
We generate $n=80$ graph signals and average the results across 50 realizations.
Fig.~\ref{fig:fnorm_minp} shows that MWGL again outperforms selected baselines and matches the theoretical results.

\subsection{Graph Learning Comparison on Molene}

\begin{figure}[hbt!]
    \centering
    \includegraphics[width=\textwidth]{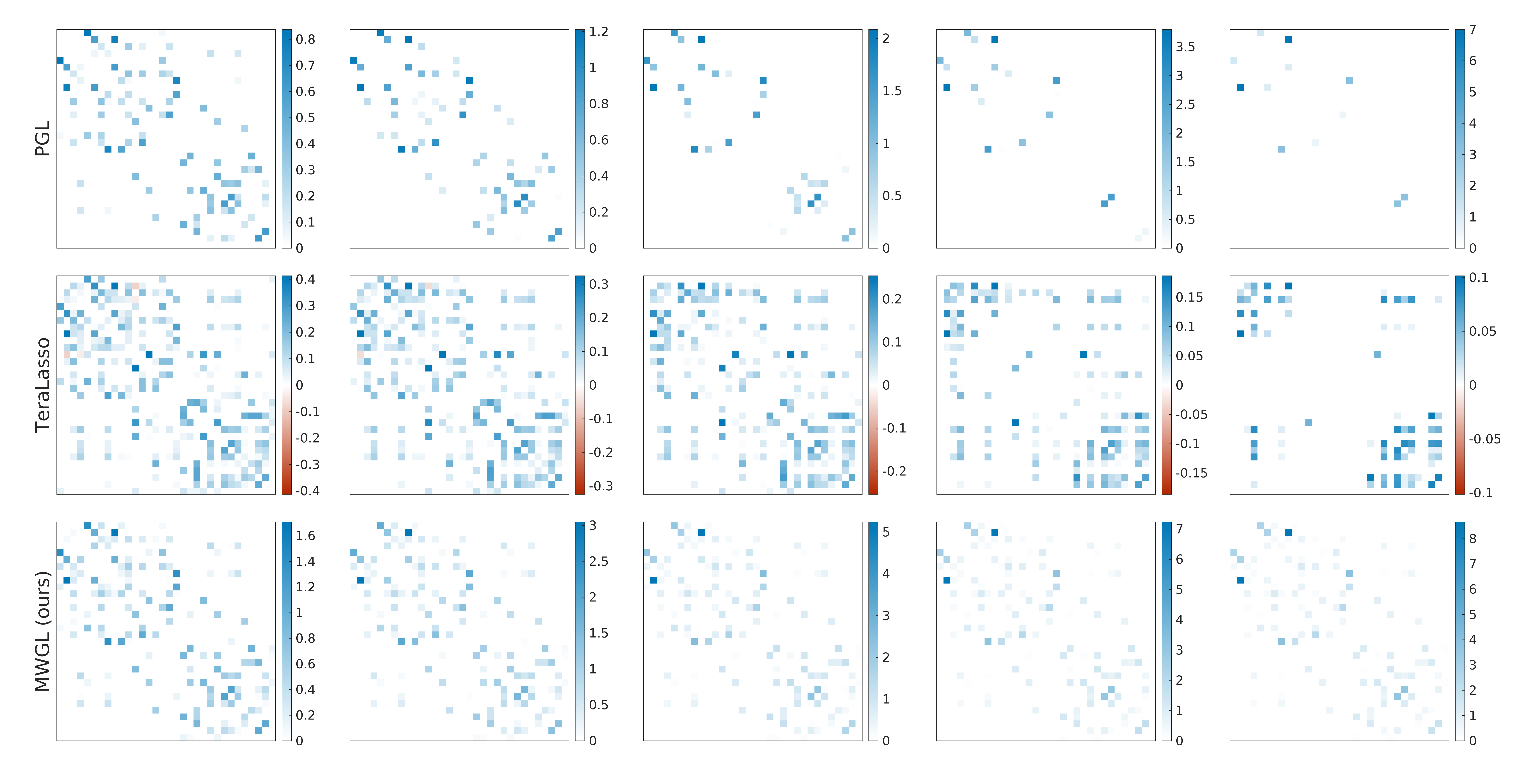}
    \caption{Comparing the learned station graph of PGL, TeraLasso, and MWGL (ours) on the Molene dataset with varying regularization.
    Laplacians are ordered with increasing sparsity from left to right.}
    \label{fig:molene_laplacian}
\end{figure}

We now compare our MWGL to PGL and TeraLasso, two methods that come close to MWGL on the synthetic data, on the Molene dataset across ranging regularization parameters.
Fig.~\ref{fig:molene_laplacian} shows the weighted adjacency matrices (negative off-diagonal precision matrices for TeraLasso) of the station graphs learned by these methods.
First, notice that TeraLasso learns few negative conditional dependencies among weather stations.
Indeed it is reasonable that the temperature of different locations does not depend negatively, which indicates that the attractive Laplacian constraints are suitable structural priors for the problem.
Also, notice that only MWGL learns connected graphs with varying regularization, and neither PGL nor TeraLasso learns connected graphs when sparsity increases.

\vfill
\end{bibunit}

\end{document}